%% file: acmmanuscript.tex
\documentclass[manuscript,screen]{acmart}

\usepackage{graphicx}
\usepackage{subcaption}

\AtBeginDocument{%
  }

\setcopyright{iw3c2w3g}
\copyrightyear{2026}
\acmYear{2026}
\usepackage{tikz}
\usetikzlibrary{calc}
\usepackage{array}
\usepackage{xcolor}

\definecolor{aucA}{HTML}{66C2A5}
\definecolor{xaucAB}{HTML}{8DA0CB}
\definecolor{xaucBA}{HTML}{E78AC3}
\definecolor{aucB}{HTML}{FC8D62}

\definecolor{parity}{HTML}{A6D854}
\definecolor{xparity}{HTML}{FFD92F}

\definecolor{xaucOne}{HTML}{80B1D3}
\definecolor{xaucZero}{HTML}{FDB462}

\colorlet{aucA}{blue!18}
\colorlet{xaucAB}{green!18}
\colorlet{xaucBA}{orange!18}
\colorlet{aucB}{red!18}

\colorlet{parity}{gray!18}
\colorlet{xparity}{yellow!25}

\colorlet{xaucOne}{cyan!18}
\colorlet{xaucZero}{violet!18}

\tikzset{
  axis label/.style={
    font=\scriptsize,
    align=center,
    text width=3.0cm
  },
  tick label/.style={
    font=\scriptsize,
    inner sep=0pt,
    outer sep=0pt
  }
}

\usepackage{booktabs}
\usepackage[table]{xcolor}
\usepackage{array}

\begin{document}

\title[Fairness Metrics based on the AUC in Clinical Predictive AI]{Fairness Metrics based on the AUC in Clinical Predictive AI}

\author{João Matos}
\orcid{0000-0002-0312-1647}
\correspondingauthor
\email{joao.matos@ndorms.ox.ac.uk}
\affiliation{%
  \institution{Centre for Statistics in Medicine, University of Oxford}
  \city{Oxford}
  \country{UK}
}

\author{Ben Van Calster}
\affiliation{%
  \institution{Department of Development and Regeneration, KU Leuven}
  \city{Leuven}
  \country{Belgium}}

\author{Richard D. Riley}
\affiliation{%
  \institution{Department of Applied Health Sciences, University of Birmingham}
  \city{Birmingham}
  \country{UK}
}
\affiliation{%
  \institution{National Institute for Health and Care Research (NIHR) Biomedical Research Centre: Birmingham, University Hospitals Birmingham NHS Foundation Trust and University of Birmingham}
  \city{Birmingham}
  \country{UK}
}

\author{Paula Dhiman}
\affiliation{%
  \institution{Centre for Statistics in Medicine, University of Oxford}
  \city{Oxford}
  \country{UK}
}

\author{Gary S. Collins}
\affiliation{%
  \institution{Department of Applied Health Sciences, University of Birmingham}
  \city{Birmingham}
  \country{UK}
}
\affiliation{%
  \institution{National Institute for Health and Care Research (NIHR) Biomedical Research Centre: Birmingham, University Hospitals Birmingham NHS Foundation Trust and University of Birmingham}
  \city{Birmingham}
  \country{UK}
}
\renewcommand{\shortauthors}{Matos et al.}

\begin{abstract}
Fairness evaluation of clinical prediction models for binary outcomes often involves fairness metrics that compare the area under the receiver operating characteristic curve (AUC, or c-statistic) across subgroups. However, the AUC has been shown to be a non-collapsible metric, meaning that its subgroup specific and overall values are not directly comparable as often implicitly assumed. This raises important questions about how existing AUC-based fairness metrics should be used and interpreted. We examine eight AUC-based fairness metrics proposed in the literature, including: within-group AUC parity; cross-group AUC (xAUC) parity; gap-to-overall AUC; equity-scaled AUC; minimum-based metrics; $\mathrm{xAUC}^1$; $\mathrm{xAUC}^0$; and $\mathrm{xAUC}^1$ or $\mathrm{xAUC}^0$ parity. We characterise their estimands, relate them to the decomposition of overall AUC (into within- and cross-group components), and critically appraise them. We then illustrate their behaviour using simulations based on the EuroSCORE II, a clinical prediction model for in-hospital mortality. Based on our formal analysis, simulations, and clinical examples, we recommend reporting an AUC matrix, containing the overall AUC and all within- and cross-group AUC components, alongside population characteristics. When fairness metrics are needed, we recommend reporting the within-group AUC parity, with cross-group AUC parity providing supplementary information but requiring caution due to its potential conflict with calibration. Metrics involving $\mathrm{xAUC}^1$ and $\mathrm{xAUC}^0$ have merit, summarising within- and cross-group AUCs into single values, but their interpretation is non-trivial, and they should be used with caution. Gap-to-overall AUC, equity-scaled AUC, and minimum-based metrics compare quantities that should not be compared and it is unclear how to operationalise them, making them not fit for purpose and generally not recommended.
\end{abstract}





\maketitle

\newpage
\section{Introduction}
\input{1_introduction}

\section{Non-collapsibility of the AUC}
\input{2_collapsibility}
\label{sec_collap}

\section{Review of AUC-based fairness metrics}
\label{sec_implications}
\input{3_metrics}

\section{Relationships between AUC-based fairness metrics}
\label{sec:relationships}

\input{4_together}

% \clearpage
\section{Critical appraisal of AUC-based fairness metrics}
\label{sec:recommendations}

\input{5_recommendations}

% \clearpage
\section{Clinical example}
\label{sec_case_study}

\input{6_euroscore}

\section{Conclusions and Recommendations}
\label{conclusion}
\input{7_conclusion}

\clearpage

\section*{Generative AI Usage Statement}
Generative AI technologies were used to assist with text review, LaTeX debugging, and Python code generation during the preparation of this manuscript. Specifically, ChatGPT Edu was employed between May 2025 and November 2026, using different versions of large language models (LLMs) over time. These tools supported manuscript preparation but were not used to draft the manuscript itself. No sensitive, proprietary, or human-subject data were shared with LLMs. All AI-generated outputs were critically reviewed and validated by the authors before incorporation, who take full responsibility for the content and integrity of this work. 

% \section*{Author Contributions}

\begin{acks}
JM is funded by a Clarendon Fund Scholarship at the University of Oxford. GSC and RDR are supported by the National Institute for Health and Care Research (NIHR) Biomedical Research Centre: Birmingham at University Hospitals Birmingham NHS Foundation Trust and the University of Birmingham; and an NIHR-MRC Better Methods Better Research Grants (MR/Z503873/1). GSC and RDR are NIHR Senior Investigators. BVC is supported by the Research Foundation – Flanders (FWO) grant G097322N, Kom Op Tegen Kanker grant 13583, and KU Leuven Internal Funds grant C24M/20/064. The funders of the study had no role in study design, data collection, data analysis, interpretation of data, writing of the report, or decision to submit the article for publication. The views expressed are those of the authors and not necessarily represent the views of the funding agencies.
\end{acks}

% \section*{Competing Interests}

% \section*{Positionality Statement}

\section*{Ethical Considerations Statement}
This study aims to improve evaluation of fairness in clinical prediction models. We highlight the limitations of existing AUC-based fairness metrics and provide recommendations for their appropriate use and interpretation. Misinterpretation or inappropriate application of these metrics could lead to misleading conclusions about model performance and fairness, potentially affecting healthcare decisions and exacerbating existing health inequalities. Our recommendations aim to support more transparent and comprehensive fairness assessments, while recognising that statistical fairness metrics alone cannot establish whether a clinical prediction model is fair or safe for deployment.

\section*{Code Availability}
The Python code used to reproduce the simulations, analyses, and figures presented in the EuroSCORE II case study (Section \ref{sec_case_study}) is publicly available as a Jupyter notebook on \href{https://colab.research.google.com/drive/12eu946PzBdCnz800UI0BwfZRziPTklON?usp=sharing}{Google Colab}. The code is distributed under the MIT Licence.

\clearpage
\bibliographystyle{ACM-Reference-Format}
\bibliography{sample-base}

\appendix
\label{app1}
\clearpage
\section{Relations between $\mathrm{xAUC^1}$, $\mathrm{xAUC^0}$ and $\mathrm{AUC}$}

\input{appendix}

\end{document}

%% file: 1_introduction.tex
Clinical prediction models using artificial intelligence (AI) promise to improve patient outcomes and reduce health care costs by informing clinical decision making and risk communication \cite{topol2019high}. When developing or evaluating a clinical prediction model, assessing its performance is essential to determine its potential value in clinical practice \cite{Van-Calster2024-jw,collins2024evaluation,matos2024clinician}.

Evaluating model performance at the target population level can conceal important differences in model behaviour across subgroups, raising \textit{fairness} concerns \cite{Oakden-Rayner2020-fn}. It is therefore common to report model performance within relevant subgroups defined by sensitive attributes such as sex or ethnicity. Although variation in model performance across subgroups is expected, understanding the nature and magnitude of differences is central to fairness evaluation \cite{mccradden2020ethical}.

A \textit{fairness metric} quantifies the extent to which a model's output does or does not discriminate, according to a particular notion of fairness, against individuals or subgroups defined by a sensitive attribute \cite{matos2025critical}. Most fairness metrics are a comparison of model performance across subgroups, for example through differences, ratios, or other summary functions of subgroup specific performance \cite{gallifant2024ethical}. Their interpretation therefore depends on the mathematical properties of the underlying performance metric. One mathematical property that is particularly relevant when comparing performance across subgroups is \textit{collapsibility} \cite{matos2026collapsibility}. A performance metric $M$ is \textit{collapsible} if its overall value can be expressed as the weighted average of its subgroup ($g$) specific values. If a metric is \textit{collapsible}, the overall value $M_{\mathrm{overall}}$ is directly related to the subgroup specific values $M_g$; when a metric is \textit{non-collapsible}, such relationship breaks down, making subgroup and overall values difficult to interpret jointly and potentially misleading fairness assessments.

One important aspect of prediction model performance for which collapsibility is particularly relevant is \textit{discrimination}, which quantifies the extent to which a model assigns higher estimated probabilities to individuals who experience the event than to those who do not \cite{Van-Calster2024-jw,Pencina2015-uu,janssens2020reflection}. The area under the receiver operating characteristic curve (AUC, also known as the c-statistic) is the most widely reported measure of discrimination for clinical prediction models for binary outcomes \cite{Van-Calster2024-jw,collins2024tripod+}. The AUC has been shown to be non-collapsible \cite{matos2026collapsibility}. Even though numerous fairness metrics that are derived directly from subgroup specific and overall AUC values have been proposed, the implications of AUC non-collapsibility for the interpretation of AUC-based fairness metrics have not been systematically examined.

In this paper, we examine AUC-based fairness metrics for clinical prediction models with binary outcomes. We review eight AUC-based fairness metrics, clarify their target estimands, relate them to the decomposition of the AUC into within- and cross-group components, and illustrate their behaviour using simulations based on the European System for Cardiac Operative Risk Evaluation (EuroSCORE) II \cite{nashef2012euroscore}. Our aim is to clarify when these metrics are informative and how they should be interpreted when evaluating fairness in clinical predictive AI.

%% file: 2_collapsibility.tex
The AUC is a non-collapsible metric \cite{matos2026collapsibility}. The overall AUC cannot, in general, be expressed as a weighted average of subgroup specific AUCs because the AUC considers all event and non-event pairs, including those from different subgroups. Consider a population partitioned into $K$ mutually exclusive subgroups. Following the notation of Kallus and Zhou \cite{Kallus2019-yl}, let $\mathrm{AUC}_{i,j}$ denote the probability that an individual with the event from subgroup $i$ is assigned, by the prediction model, a higher estimated probability than an individual without the event from subgroup $j$. When $i=j$, $\mathrm{AUC}_{i,i}$ corresponds to the usual within-group AUC, whereas for $i\neq j$, $\mathrm{AUC}_{i,j}$ is the cross-group AUC (xAUC). The overall AUC can therefore be decomposed \cite{matos2026collapsibility} as:
\begin{align}
\mathrm{AUC}_{\mathrm{overall}}
&=
\sum_{i,j=1}^{K}
\frac{p_1^i}{p_1}
\frac{p_0^j}{p_0}
\,
\mathrm{AUC}_{i,j},
\label{eq:auc_decomposition_general}
\end{align}
where $p_1^i$ and $p_0^i$ denote the numbers (or proportions) of event and non-event observations in subgroup $i$, respectively, and $p_1=\sum_i p_1^i$ and $p_0=\sum_i p_0^i$ denote the corresponding totals. For two subgroups, $A$ and $B$, Equation~\eqref{eq:auc_decomposition_general} reduces to
\begin{align}
\mathrm{AUC}_{\mathrm{overall}}
&=
\frac{p_1^A p_0^A}{p_1 p_0}\,
\mathrm{AUC}_{A}
+
\frac{p_1^A p_0^B}{p_1 p_0}\,
\mathrm{xAUC}_{A,B}
+
\frac{p_1^B p_0^A}{p_1 p_0}\,
\mathrm{xAUC}_{B,A}
+
\frac{p_1^B p_0^B}{p_1 p_0}\,
\mathrm{AUC}_{B}
\label{eq:auc_decomposition_two_groups}
\end{align}

Equation~\eqref{eq:auc_decomposition_general} shows that the overall AUC is a weighted combination of within- and cross-group AUCs. Consequently, the overall AUC is generally not determined by the within-group AUCs alone, hence why it is non-collapsible \cite{matos2026collapsibility}. 

%% file: 3_metrics.tex
We now briefly examine eight AUC-based fairness metrics that have been proposed in the literature and summarised by \cite{matos2025critical}, through the lens of the AUC decomposition (presented in Section \ref{sec_collap}). These metrics are based on different components of the AUC and therefore capture different aspects of model discrimination. For consistency with the notation introduced above, we write \(\mathrm{AUC}\) for the overall AUC, \(\mathrm{AUC}_{g}\) for the within-group AUC in subgroup \(g\), and \(\mathrm{xAUC}_{i,j}\) for the cross-group AUC comparing individuals with the event from subgroup \(i\) against individuals without the event from subgroup \(j\). We let \(\hat{p}_y^{g}\) denote the estimated probability assigned by a prediction model for binary outcome \(y\in\{0,1\}\) for an arbitrary individual in subgroup \(g\in\{A,B\}\).

We only consider two subgroups (\(A\) and \(B\)) for simplicity. However, the metrics presented in this section can be extended to \(n>2\) subgroups by computing pairwise differences across all \(\binom{n}{2}=\frac{n(n-1)}{2}\) distinct subgroup pairs. 

A fairness metric $M$ may be operationalised using a pre-specified threshold $\theta$ to determine whether a model satisfies the fairness notion at hand. For a metric $M$ where larger values indicate greater fairness, this may be defined as \(M \geq \theta\). Conversely, for a metric where smaller values indicate greater fairness, this may be defined as \(M \leq \theta\). The choice of $\theta$ depends on the metric and application. For example, a parity-based metric, where smaller values indicate greater fairness, may use $\theta=0.05$ as the maximum acceptable difference.

\subsection{Within-group AUC parity}
Within-group AUC parity calculates the AUC within each subgroup and then estimates their absolute difference \cite{cheng2021can}:
\begin{align}
\mathrm{AUC}_{\mathrm{parity}}
=
\left|
\mathrm{AUC}_{A}
-
\mathrm{AUC}_{B}
\right|.
\label{auc_parity}
\end{align}

A multiplicative variant has also been proposed \cite{byrd2023validation},
\begin{align}
\mathrm{AUC}_{\mathrm{ratio}}
=
\frac{\mathrm{AUC}_{A}}
{\mathrm{AUC}_{B}}.
\label{auc_ratio}
\end{align}

Both metrics quantify asymmetries in within-group discrimination only. For the difference-based metric (equation \ref{auc_parity}), the ``optimal value'' is 0, whereas for the ratio-based one (equation \ref{auc_ratio}) a value of 1 indicates equal performance.

\subsection{Cross-group AUC (xAUC) parity}
Cross-group AUC (i.e., ranking of individuals with an event from one subgroup versus individuals without an event from the other subgroup) is calculated for each subgroup, the absolute difference is calculated across subgroups \cite{Kallus2019-yl}:
\begin{align}
\mathrm{xAUC}_{\mathrm{parity}}
=
\left|
\mathrm{xAUC}_{A,B}
-
\mathrm{xAUC}_{B,A}
\right|.
\end{align}

A value of ``0'' means that the model is fair with respect to this metric (i.e., no cross-group discrimination asymmetry).

\subsection{Gap to overall AUC}
It has been proposed to compare within-group AUC with the overall AUC across subgroups \cite{albuquerque2024evaluating}. This metric calculates the AUC in the overall sample, in the subgroup, and then takes the absolute difference:

\begin{align}
\mathrm{AUC}_{\mathrm{gap}}
=
\left|
\mathrm{AUC}
-
\mathrm{AUC}_{g}
\right|.
\end{align}

A value of ``0'' means that the model is fair with respect to this metric.

\subsection{Equity-scaled AUC}
Luo et al.~\cite{Luo2024-uo} proposed penalising the overall AUC according to the within-group AUC parity:
\begin{align}
\mathrm{AUC_{ES}}
=
\frac{\mathrm{AUC}}
{1+\mathrm{AUC}_{\mathrm{parity}}}.
\end{align}

When $\mathrm{AUC_{ES}} = \mathrm{AUC}$ (i.e., $\mathrm{AUC_{parity}} = 0$), the model is considered fair with respect to this metric.

\subsection{Minimum-based metrics}
Minimum-based metrics can evaluate discrimination according to the worst-performing subgroup \cite{diana2021minimax,mittelstadt2023unfairnessfairmachinelearning}, for example:
\begin{align}
\mathrm{AUC}_{\min}
=
\min
\left\{
\mathrm{AUC}_{A},
\mathrm{AUC}_{B}
\right\}.
\end{align}

Analogous definitions may be constructed using other AUC components. A model is considered fair when the minimum value of a metric of interest (e.g., $AUC_g$) is above a pre-specified threshold (e.g., $\min \left\{AUC_g \right\} \geq 0.8$).

\subsection{\boldmath\texorpdfstring{$\mathrm{xAUC}^1$}{xAUC1}}
Also referred to as ``Pairwise Ranking Fairness'' (PRF), $\mathrm{xAUC}^1$ quantifies the probability that an individual with the event from subgroup \(g\) is ranked above a randomly selected individual without the event from the population \cite{Kallus2019-yl,Cui2023-lo}:
\begin{align}
\mathrm{xAUC}^1_A
&=
P(\hat p_1^A>\hat p_0),
\\
\mathrm{xAUC}^1_B
&=
P(\hat p_1^B>\hat p_0).
\end{align}

A higher $\mathrm{xAUC}^1_g$ indicates that individuals with the event from subgroup $g$ are more likely to be assigned higher estimated probabilities than individuals without the event in the overall population. A value of $0.5$ corresponds to chance-level ranking, while values approaching $1$ indicate better discrimination. Fairness is achieved when $\mathrm{xAUC}^1_A=\mathrm{xAUC}^1_B$, indicating that individuals with the event are ranked relative to the non-event population equally well across subgroups. 

\subsection{\boldmath\texorpdfstring{$\mathrm{xAUC}^0$}{xAUC0}}
Also referred to as ``Subgroup AUC'' (sAUC), $\mathrm{xAUC}^0$ quantifies the probability that a randomly selected individual without the event from subgroup \(g\) is ranked below any individual with the event from the population \cite{Kallus2019-yl,Meissen2024-iu}:
\begin{align}
\mathrm{xAUC}^0_A
&=
P(\hat p_1>\hat p_0^A),
\\
\mathrm{xAUC}^0_B
&=
P(\hat p_1>\hat p_0^B).
\end{align}

Similarly to $\mathrm{xAUC}^1$ (above), fairness is achieved when $\mathrm{xAUC}^0_A=\mathrm{xAUC}^0_B$, indicating that individuals without the event are ranked relative to the event population equally well across subgroups.

\subsection{\boldmath\texorpdfstring{$\mathrm{xAUC}^1$ and $\mathrm{xAUC}^0$ parity}{xAUC1 and xAUC0 parity}}
These metrics estimate absolute differences in $\mathrm{xAUC}^1$ or $\mathrm{xAUC}^0$, respectively, across subgroups \cite{Meissen2024-iu,Cui2023-lo}:
\begin{align}
\mathrm{xAUC}^1_{\mathrm{parity}}
=
\left|
\mathrm{xAUC}^1_A
-
\mathrm{xAUC}^1_B
\right|,
\end{align}
\begin{align}
\mathrm{xAUC}^0_{\mathrm{parity}}
=
\left|
\mathrm{xAUC}^0_A
-
\mathrm{xAUC}^0_B
\right|
\end{align}

Fairness is achieved when $\mathrm{xAUC}^1_{\mathrm{parity}} =0 $ or $\mathrm{xAUC}^0_{\mathrm{parity}} = 0$.

%% file: 4_together.tex
Several of the metrics presented above are either related to each other or to the overall AUC. For example, $\mathrm{xAUC}^1$ and $\mathrm{xAUC}^0$ can be expressed in terms of within-group and cross-group discrimination. For two subgroups $A$ and $B$,

\begin{align}
\mathrm{xAUC}^1_{A}
&=
\frac{p_0^A}{p_0}\,
\mathrm{AUC}_{A}
+
\frac{p_0^B}{p_0}\,
\mathrm{xAUC}_{A,B},
\\
\mathrm{xAUC}^1_{B}
&=
\frac{p_0^A}{p_0}\,
\mathrm{xAUC}_{B,A}
+
\frac{p_0^B}{p_0}\,
\mathrm{AUC}_{B},
\\
\mathrm{xAUC}^0_{A}
&=
\frac{p_1^A}{p_1}\,
\mathrm{AUC}_{A}
+
\frac{p_1^B}{p_1}\,
\mathrm{xAUC}_{B,A},
\\
\mathrm{xAUC}^0_{B}
&=
\frac{p_1^A}{p_1}\,
\mathrm{xAUC}_{A,B}
+
\frac{p_1^B}{p_1}\,
\mathrm{AUC}_{B}.
\end{align}

These four quantities can in turn form a linear decomposition of the overall AUC,

\begin{align}
\mathrm{AUC}
=
\frac{p_1^A}{p_1}
\mathrm{xAUC}^1_A
+
\frac{p_1^B}{p_1}
\mathrm{xAUC}^1_B
=
\frac{p_0^A}{p_0}
\mathrm{xAUC}^0_A
+
\frac{p_0^B}{p_0}
\mathrm{xAUC}^0_B,
\end{align}

providing an alternative representation of the overall model discrimination. Full proofs can be found in the Appendix.

Unlike the commonly reported within-group AUC, $\mathrm{xAUC}^1$ and $\mathrm{xAUC}^0$ incorporate both within- and cross-group discrimination. An alternative way to interpret these decompositions is through a part-whole perspective. $\mathrm{xAUC}^1_A$ and $\mathrm{xAUC}^1_B$ (and analogously $\mathrm{xAUC}^0_A$ and $\mathrm{xAUC}^0_B$) form mutually exclusive and collectively exhaustive contributions to the overall AUC, a property that conventional within-group AUCs lack since they omit the cross-group components.

Figure~\ref{figure_metrics} summarises AUC components and AUC-based fairness metrics using AUC matrices. For a binary sensitive attribute, an \emph{AUC Matrix} partitions all pairwise comparisons according to the subgroup membership of the individual with the event (rows) and the individual without the event (columns) \cite{chang2022disparate}. Each cell represents the probability that a randomly selected individual with an event from one subgroup has a higher estimated probability than a randomly selected individual without an event from the same or the other subgroup. The diagonal cells correspond to the within-group AUCs ($\mathrm{AUC}_A$ and $\mathrm{AUC}_B$) and the off-diagonal cells to the cross-group AUCs ($\mathrm{xAUC}_{A,B}$ and $\mathrm{xAUC}_{B,A}$).

\input{figure_metrics}

% \paragraph{xAUC_parity}

% \begin{align}
% \mathrm{xAUC}^1_A-\mathrm{xAUC}^1_B
% &=
% \left[
% \frac{p_0^A}{p_0}\mathrm{AUC}_A
% +
% \frac{p_0^B}{p_0}\mathrm{xAUC}_{A,B}
% \right]
% -
% \left[
% \frac{p_0^A}{p_0}\mathrm{xAUC}_{B,A}
% +
% \frac{p_0^B}{p_0}\mathrm{AUC}_B
% \right]
% \\[4pt]
% &=
% \frac{p_0^A}{p_0}
% \underbrace{
% \left(
% \mathrm{AUC}_A-\mathrm{AUC}_B
% \right)
% }_{\text{± within-group parity}}
% +
% \frac{p_0^B}{p_0}
% \underbrace{
% \left(
% \mathrm{xAUC}_{A,B}-\mathrm{xAUC}_{B,A}
% \right)
% }_{\text{± cross-group parity}}
% +
% \underbrace{
% \left(
% \frac{p_0^A}{p_0}-\frac{p_0^B}{p_0}
% \right)
% \left(
% \mathrm{AUC}_B-\mathrm{xAUC}_{B,A}
% \right)
% }_{\text{``weight correction term?''}}.
% \end{align}

% For the special case in which $\frac{p_0^A}{p_0}=\frac{p_0^B}{p_0}=\frac{1}{2}$ the weight correction term is zero. 

%% file: figure_metrics.tex
% ------------------------------------------------------------
% Preamble requirements
% ------------------------------------------------------------
% \usepackage{tikz}
% \usetikzlibrary{calc}
% \usepackage{xcolor}

% ------------------------------------------------------------
% Colour definitions
% ------------------------------------------------------------
\colorlet{aucA}{blue!18}
\colorlet{xaucAB}{green!18}
\colorlet{xaucBA}{orange!35}
\colorlet{aucB}{red!18}
\colorlet{parity}{gray!18}
\colorlet{xparity}{yellow!25}
\colorlet{xaucOne}{cyan!18}
\colorlet{xaucZero}{violet!18}
\colorlet{xaucOneAlt}{brown!25}
\colorlet{xaucZeroAlt}{pink!22}

% ------------------------------------------------------------
% Styles (only needed if not already defined elsewhere in your doc —
% delete this block if you already have "tick label" / "axis label" styles)
% ------------------------------------------------------------
\tikzset{
  tick label/.style={font=\scriptsize},
  axis label/.style={font=\scriptsize, align=center}
}

% ------------------------------------------------------------
% Fixed row geometry — SAME for every row, regardless of how long
% that row's legend text is. This is what keeps the matrices
% aligned: \centering centers each tikzpicture based on its
% bounding box, so if the box width/left-edge varied row to row
% (because legend text lengths differ), the matrix would shift
% left/right by a different amount in each row. Pinning the box
% here removes that variation.
% ------------------------------------------------------------
\newlength{\AUCRowLeft}
\newlength{\AUCRowRight}
\setlength{\AUCRowLeft}{-3cm}   % clears the rotated left-hand axis label
\setlength{\AUCRowRight}{10cm}  % clears the widest legend text (8.7cm) + \legX (3.2cm)

% ------------------------------------------------------------
% Draws the 2x2 matrix + row/column/axis labels at the CURRENT origin.
% Does not open its own tikzpicture — meant to be called inside one.
% ------------------------------------------------------------
\newcommand{\AUCMatrixDraw}[8]{%
  \pgfmathsetlengthmacro{\cell}{1.25cm}
  \coordinate (O) at (0,0);
  \useasboundingbox (\AUCRowLeft,0.5cm) rectangle (\AUCRowRight,-3.0cm);
  % fills
  \fill[#1] (O) rectangle ($(O)+(\cell,-\cell)$);
  \fill[#2] ($(O)+(\cell,0)$) rectangle ($(O)+(2*\cell,-\cell)$);
  \fill[#3] ($(O)+(0,-\cell)$) rectangle ($(O)+(\cell,-2*\cell)$);
  \fill[#4] ($(O)+(\cell,-\cell)$) rectangle ($(O)+(2*\cell,-2*\cell)$);
  % grid lines
  \draw[line width=0.45pt] (O) rectangle ($(O)+(2*\cell,-2*\cell)$);
  \draw[line width=0.45pt] ($(O)+(\cell,0)$) -- ($(O)+(\cell,-2*\cell)$);
  \draw[line width=0.45pt] ($(O)+(0,-\cell)$) -- ($(O)+(2*\cell,-\cell)$);
  % cell texts
  \node[font=\scriptsize] at ($(O)+(.5*\cell,-.5*\cell)$) {#5};
  \node[font=\scriptsize] at ($(O)+(1.5*\cell,-.5*\cell)$) {#6};
  \node[font=\scriptsize] at ($(O)+(.5*\cell,-1.5*\cell)$) {#7};
  \node[font=\scriptsize] at ($(O)+(1.5*\cell,-1.5*\cell)$) {#8};
  % row labels
  \node[tick label, anchor=east] at ($(O)+(-0.10cm,-0.50*\cell)$) {A};
  \node[tick label, anchor=east] at ($(O)+(-0.10cm,-1.50*\cell)$) {B};
  % column labels
  \node[tick label, anchor=north] at ($(O)+(0.50*\cell,-2*\cell-0.08cm)$) {A};
  \node[tick label, anchor=north] at ($(O)+(1.50*\cell,-2*\cell-0.08cm)$) {B};
  % axis labels
  \node[axis label, rotate=90, anchor=center]
    at ($(O)+(-0.86cm,-1.0*\cell)$)
    {Individuals with an event\\from subgroup};
  \node[axis label, anchor=north]
    at ($(O)+(1.0*\cell,-2*\cell-0.48cm)$)
    {Individuals without an event\\from subgroup};
}

% ------------------------------------------------------------
% Places one legend entry (colour swatch + text) at a given y,
% in the SAME coordinate system as the matrix. \legX is the x
% position where the legend column starts (defined per row below).
% Because the node anchor is "west", its y-coordinate is the
% vertical CENTER of the (possibly multi-line) node, so the
% swatch — drawn centered on that same y — always lines up with
% the text, no matter how many lines the equation wraps to.
% ------------------------------------------------------------
\newcommand{\LegendItem}[4]{%
  % #1 fill colour, #2 y position, #3 text width, #4 content
  \node[anchor=west, align=left, text width=#3] (li) at (\legX,#2) {#4};
  \filldraw[fill=#1, draw=black] ($(\legX,#2)+(-0.25cm,-0.09cm)$) rectangle ++(0.18,0.18);
}

% ------------------------------------------------------------
% Figure
% ------------------------------------------------------------
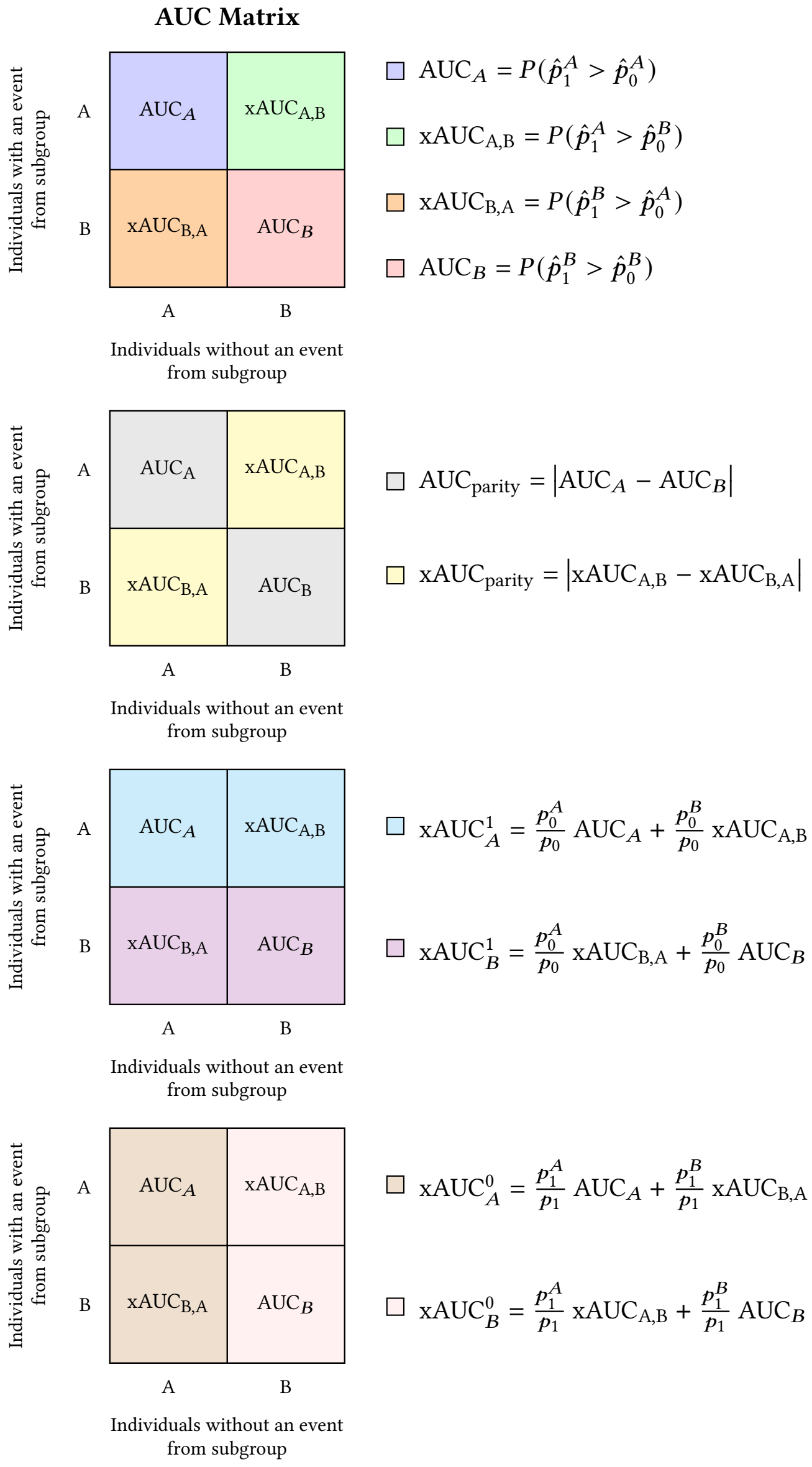
\begin{figure}[ht]
\centering
\small
\vspace{1em}
% ------------------------------------------------------------
% Row 1 — 4 legend entries
% ------------------------------------------------------------
\begin{tikzpicture}[baseline=(current bounding box.north)]
  \AUCMatrixDraw
    {aucA}{xaucAB}{xaucBA}{aucB}
    {$\mathrm{AUC}_A$}{$\mathrm{xAUC}_{\mathrm{A,B}}$}{$\mathrm{xAUC}_{\mathrm{B,A}}$}{$\mathrm{AUC}_B$}

  \node[font=\small\bfseries] at (1.25cm,0.38cm) {AUC Matrix};
  
  \def\legX{3.2cm}
  \LegendItem{aucA}{-0.20}{6cm}{$\mathrm{AUC}_A = P(\hat{p}_1^A > \hat{p}_0^A)$}
  \LegendItem{xaucAB}{-0.90}{6cm}{$\mathrm{xAUC}_{\mathrm{A,B}} = P(\hat{p}_1^A > \hat{p}_0^B)$}
  \LegendItem{xaucBA}{-1.60}{6cm}{$\mathrm{xAUC}_{\mathrm{B,A}} = P(\hat{p}_1^B > \hat{p}_0^A)$}
  \LegendItem{aucB}{-2.30}{6cm}{$\mathrm{AUC}_B = P(\hat{p}_1^B > \hat{p}_0^B)$}
\end{tikzpicture}

\vspace{1em}

% ------------------------------------------------------------
% Row 2 — 2 legend entries (single-line equations)
% ------------------------------------------------------------
\begin{tikzpicture}[baseline=(current bounding box.north)]
  \AUCMatrixDraw
    {parity}{xparity}{xparity}{parity}
    {$\mathrm{AUC}_{\mathrm{A}}$}{$\mathrm{xAUC}_{\mathrm{A,B}}$}{$\mathrm{xAUC}_{\mathrm{B,A}}$}{$\mathrm{AUC}_{\mathrm{B}}$}
  \def\legX{3.2cm}
  \LegendItem{parity}{-0.75}{7.8cm}{$\mathrm{AUC}_{\mathrm{parity}} = \big|\mathrm{AUC}_A - \mathrm{AUC}_B\big|$}
  \LegendItem{xparity}{-1.75}{7.8cm}{$\mathrm{xAUC}_{\mathrm{parity}} = \big|\mathrm{xAUC}_{\mathrm{A,B}} - \mathrm{xAUC}_{\mathrm{B,A}}\big|$}
\end{tikzpicture}

\vspace{1em}

% ------------------------------------------------------------
% Row 3 — 2 legend entries (stacked-fraction equations, extra room)
% ------------------------------------------------------------
\begin{tikzpicture}[baseline=(current bounding box.north)]
  \AUCMatrixDraw
    {xaucOne}{xaucOne}{xaucZero}{xaucZero}
    {$\mathrm{AUC}_A$}{$\mathrm{xAUC}_{\mathrm{A,B}}$}{$\mathrm{xAUC}_{\mathrm{B,A}}$}{$\mathrm{AUC}_B$}
  \def\legX{3.2cm}
  \LegendItem{xaucOne}{-0.60}{8.7cm}{%
    $\mathrm{xAUC}^1_{A}
    =
    \frac{p_0^A}{p_0}\,\mathrm{AUC}_{A}
    +
    \frac{p_0^B}{p_0}\,\mathrm{xAUC}_{\mathrm{A,B}}$}
  \LegendItem{xaucZero}{-1.90}{8.7cm}{%
    $\mathrm{xAUC}^1_{B}
    =
    \frac{p_0^A}{p_0}\,\mathrm{xAUC}_{\mathrm{B,A}}
    +
    \frac{p_0^B}{p_0}\,\mathrm{AUC}_{B}$}
\end{tikzpicture}

\vspace{1em}

% ------------------------------------------------------------
% Row 4 — 2 legend entries (stacked-fraction equations, extra room)
% ------------------------------------------------------------
\begin{tikzpicture}[baseline=(current bounding box.north)]
  \AUCMatrixDraw
    {xaucOneAlt}{xaucZeroAlt}{xaucOneAlt}{xaucZeroAlt}
    {$\mathrm{AUC}_A$}{$\mathrm{xAUC}_{\mathrm{A,B}}$}{$\mathrm{xAUC}_{\mathrm{B,A}}$}{$\mathrm{AUC}_B$}
  \def\legX{3.2cm}
  \LegendItem{xaucOneAlt}{-0.60}{8.7cm}{%
    $\mathrm{xAUC}^0_{A}
    =
    \frac{p_1^A}{p_1}\,\mathrm{AUC}_{A}
    +
    \frac{p_1^B}{p_1}\,\mathrm{xAUC}_{\mathrm{B,A}}$}
  \LegendItem{xaucZeroAlt}{-1.90}{8.7cm}{%
    $\mathrm{xAUC}^0_{B}
    =
    \frac{p_1^A}{p_1}\,\mathrm{xAUC}_{\mathrm{A,B}}
    +
    \frac{p_1^B}{p_1}\,\mathrm{AUC}_{B}$}
\end{tikzpicture}

\vspace{2em}

\caption{Relationship between different AUC metrics (for binary outcome \(y\in\{0,1\}\) and assuming two subgroups \(A\) and \(B\)).}
\label{figure_metrics}
\end{figure}

%% file: 5_recommendations.tex
In this section, we critically appraise each of the metrics in Section \ref{sec_implications}, in light of their formalisation and relationships with one another. Table~\ref{tab:fairness-metric-summary} summarises the estimand targeted by each metric, our critical appraisal, and the corresponding recommendation. Recommendations are colour coded as \textcolor{green!60!black}{recommended}, \textcolor{orange}{use with caution}, and \textcolor{red!70!black}{generally not recommended}. The table is intended as a practical guide for selecting AUC-based fairness metrics that align with an intended fairness objective whilst avoiding misleading interpretations. In general, we recommend reporting the overall AUC and the complete \textit{AUC matrix} in Figure~\ref{figure_metrics}, which makes all within-group and cross-group AUC components explicit.

\subsection{Within-group AUC parity}
Within-group AUC parity \cite{cheng2021can} compares the AUC calculated within each subgroup. It is appropriate when the fairness objective is to assess whether the model discriminates equally well within each subgroup. As demonstrated by the decomposition of the AUC, within-group AUCs ignore cross-group components, which may differ substantially even when subgroup specific AUCs are identical. Consequently, similar values of $\mathrm{AUC}_A$ and $\mathrm{AUC}_B$ may coexist with pronounced asymmetries in $\mathrm{xAUC}_{A,B}$ and $\mathrm{xAUC}_{B,A}$, as well as with apparently paradoxical behaviour of the overall AUC due to non-collapsibility \cite{matos2026collapsibility}. Conversely, substantial differences in within-group AUCs do not necessarily imply disparities in cross-group AUCs. This metric is \textcolor{green!60!black}{recommended} as the primary fairness metric for discrimination performance and should be reported alongside overall AUC. As the within-group AUC parity does not capture cross-group ranking disparities, the xAUC parity can provide important additional information, with cautious interpretation.

\subsection{Cross-group AUC (xAUC) parity}
Cross-group AUC (xAUC) parity is relevant when fairness concerns how individuals from different groups are ranked relative to one another (e.g., patients are ranked based on estimated probabilities or a binary decision is made after applying a single threshold). This metric provides information beyond the within-group AUC parity, which is important to comprehensively evaluate fairness in discrimination performance. However, as discussed by \cite{matos2026collapsibility}, xAUC parity may be influenced by differences in outcome prevalence or calibration across subgroups. Consequently, improvements in xAUC parity may come at the expense of calibration disparity. Calibration has been recommended as a key domain of model performance \cite{Van-Calster2024-jw} and fairness evaluation \cite{matos2025critical,matos2026jama}, unless only patient ranking matters \cite{Kallus2019-yl}. The xAUC parity should be \textcolor{orange}{used with caution} and reported alongside within-group AUC parity and overall AUC. Interpreting this metric in isolation may be misleading because of its interplay with calibration and case-mix differences across subgroups.

% Cross-group AUC parity evaluates disparities in cross-group discrimination performance. This is appropriate when fairness concerns how individuals from different groups are ranked relative to one another. Similar to within-group AUC parity, it captures only one aspect of discrimination (cross-group). This type of metric is \textcolor{green!60!black}{recommended} when reported together with within-group AUC parity (and overall AUC). Reporting this metric on its own would imply that xAUC parity is the primary fairness objective (which is unlikely to be the case), and interpretation would be non-trivial.

\subsection{Gap to overall AUC}
The difference between the overall AUC and a subgroup specific AUC compares quantities that estimate fundamentally different aspects of discrimination. Subgroup specific AUC values quantify only within-group discrimination, whereas the overall AUC combines within-group and cross-group components. As these quantities are generally not directly comparable, this type of metric is \textcolor{red!70!black}{generally not recommended} for fairness checks.

\subsection{Equity-Scaled AUC}
Equity-Scaled AUC divides overall AUC by within-group AUC parity. The numerator (overall AUC) depends on both within- and cross-group AUCs, whereas the denominator depends only on within-group AUC parity. The metric implicitly assumes that fairness (``equity'') is captured by within-group parity alone, ignoring xAUCs. If used in isolation, this metric is difficult to interpret and susceptible to manipulation. This metric is \textcolor{red!70!black}{generally not recommended}.

\subsection{Minimum-based fairness metrics}
Minimum-based metrics focus exclusively on the worst-performing subgroup and aim to align with fairness definitions based on maximising the minimum subgroup performance \cite{mittelstadt2023unfairnessfairmachinelearning}. However, it is unclear how to interpret when reported in isolation, or how to act upon disparities flagged by this metric alone. This type of metric is \textcolor{red!70!black}{generally not recommended}. 

\subsection{\boldmath$\mathrm{xAUC}^1$}
$\mathrm{xAUC}^1_g$ measures discrimination for individuals with the event ($Y=1$) in subgroup $g$, by comparing their estimated probabilities against individuals without the event in the overall population. $\mathrm{xAUC}^1_g$ incorporates both within- and cross-group discrimination through a linear decomposition (Section~\ref{sec:relationships}). Consequently, it is less susceptible to Simpson-type paradoxes arising from ignoring cross-group AUCs. Whether this metric qualifies as a fairness metric on its own, or as an ancillary metric for fairness evaluation, is unclear. For example, for two subgroups, a comparison between $\mathrm{xAUC}^1_A$ and $\mathrm{xAUC}^1_B$ would be necessary to draw any fairness conclusion, and $\mathrm{xAUC}^1_A$ or $\mathrm{xAUC}^1_B$ do not say much about fairness on their own. This metric should be \textcolor{orange}{used with caution} for each subgroup, when fairness is viewed from the perspective of individuals who do experience the event, and both within- and cross-group discrimination are relevant.

\subsection{\boldmath$\mathrm{xAUC}^0$}
$\mathrm{xAUC}^0$ measures discrimination for individuals without the event ($Y=0$) in subgroup $g$. Like $\mathrm{xAUC}^1$, it combines within- and cross-group discrimination into a linear decomposition of the overall AUC. This metric should be \textcolor{orange}{used with caution} when fairness is viewed from the perspective of individuals who do not experience the event, and both within- and cross-group discrimination are relevant. 

\subsection{\boldmath$\mathrm{xAUC}^1$ and \boldmath$\mathrm{xAUC}^0$ Parity}
These metrics should be \textcolor{orange}{used with caution} whenever the use of $\mathrm{xAUC}^1$ or $\mathrm{xAUC}^0$ is deemed justifiable and relevant. Fairness conclusions based on the $\mathrm{xAUC}^1_{\mathrm{parity}}$ may be particularly relevant in positively polar decisions (i.e., patients benefit from \textit{selection}, e.g., being selected for an organ transplant). An eligible patient from subgroup $g$ should always be ranked (by the model) \textit{above} an ineligible patient from any subgroup \cite{Paulus_Kent_2020}, regardless of subgroup membership. Contrary to $\mathrm{xAUC}^1_{\mathrm{parity}}$, $\mathrm{xAUC}^0_{\mathrm{parity}}$ is particularly relevant in \textit{negatively} polar decisions (i.e, patients benefit from \textit{exclusion}, e.g., quarantining patients at high infectious risk). An ineligible patient from subgroup $g$ should always be ranked (by the model) \textit{below} an eligible patient from any subgroup \cite{Paulus_Kent_2020}, regardless of subgroup membership.

\input{recs_table}

%% file: recs_table.tex
% Table
\begin{table}[t]
\centering
\footnotesize
\caption{Summary of AUC-based fairness metrics, their meanings, critical appraisal, and recommendations.}
\label{tab:fairness-metric-summary}

\renewcommand{\arraystretch}{1.25}
\begin{tabular}{>{\raggedright\arraybackslash}p{1.8cm}
                >{\raggedright\arraybackslash}p{2.5cm}
                >{\raggedright\arraybackslash}p{3cm}
                >{\raggedright\arraybackslash}p{6.5cm}}
\toprule
\textbf{Metric} &
\textbf{Formula} &
\textbf{Meaning} &
\textbf{Critical appraisal and recommendation} \\
\midrule

Within-group AUC parity \cite{cheng2021can,byrd2023validation}& \(\displaystyle \left|\mathrm{AUC}_{A}-\mathrm{AUC}_{B}\right|\)
& Absolute difference in within-group discrimination & \textcolor{green!60!black}{\textbf{Recommended.}} Appropriate as the primary fairness metric for discrimination performance. Should be reported alongside overall AUC. Can be used on its own, with cross-group AUC parity providing important additional information for a more comprehensive evaluation of fairness in model discrimination. \\\\

% Cross-group AUC parity \cite{Kallus2019-yl} &  \(\displaystyle \left|\mathrm{xAUC}_{A,B}-\mathrm{xAUC}_{B,A}\right| \) & Absolute difference in cross-group discrimination &
% \textcolor{green!60!black}{\textbf{Recommended.}} Should be reported together with within-group AUC parity. Interpretation in isolation is not very intuitive but it is appropriate when fairness concerns primarily involve cross-group rankings. \\\\

Cross-group AUC parity \cite{Kallus2019-yl} &  \(\displaystyle \left|\mathrm{xAUC}_{A,B}-\mathrm{xAUC}_{B,A}\right| \) & Absolute difference in cross-group discrimination &
\textcolor{orange}{\textbf{Use with caution.}} Provides important additional information on xAUC disparities, but it may be influenced by differences in outcome prevalence and calibration across subgroups. Should be reported alongside within-group AUC parity and overall AUC. Interpretation in isolation may be misleading. \\\\

Gap to overall AUC \cite{albuquerque2024evaluating} & \(\displaystyle \left|\mathrm{AUC}-\mathrm{AUC}_{g}\right|\)& Absolute difference between overall and within-group discrimination &
\textcolor{red!70!black}{\textbf{Not recommended.}} Compares quantities estimating fundamentally different aspects of discrimination. Interpretation is generally convoluted. \\\\

``Equity-Scaled'' AUC \cite{Luo2024-uo}& \(\displaystyle
\frac{\mathrm{AUC}}{1+\mathrm{AUC}_{\mathrm{parity}}}\) & Overall discrimination penalised by within-group discrimination parity &
\textcolor{red!70!black}{\textbf{Not recommended.}} Combines quantities targeting different aspects of discrimination. It is difficult to interpret and offers limited added value when other metrics are already reported. \\\\

Minimum-based metrics \cite{diana2021minimax,mittelstadt2023unfairnessfairmachinelearning} & \(\displaystyle \min\{\mathrm{AUC}_{A},\mathrm{AUC}_{B}\}\)
& Worst-performing subgroup discrimination & \textcolor{red!70!black}{\textbf{Not recommended.}} Aligns with fairness objectives based on maximising worst-case subgroup performance but it is unclear how to interpret on its own or act upon potential disparities. \\\\

$\mathrm{xAUC}^1$ \cite{Kallus2019-yl,Cui2023-lo} & \(\displaystyle P(\hat p_1^g>\hat p_0)\)
& Probability that an individual in subgroup $g$ with the event is ranked above any individual without the event & \textcolor{orange}{\textbf{Use with caution.}} Interpretation is non-trivial. Combines within-group and cross-group discrimination for a subgroup $g$ in a single metric, focusing on individuals with the event. Contains more information than the within-group AUC alone but it is still a summary rather than the full AUC matrix. \\\\

$\mathrm{xAUC}^0$ \cite{Kallus2019-yl,Meissen2024-iu} & \(\displaystyle P(\hat p_1>\hat p_0^g)\) & Probability that an individual in subgroup $g$ without the event is ranked below any individual with the event & \textcolor{orange}{\textbf{Use with caution.}} Interpretation is non-trivial. Combines within-group and cross-group discrimination for a subgroup $g$ in a single metric, focusing on individuals without the event. Contains more information than the within-group AUC alone but it does not replace the full AUC matrix.\\\\

$\mathrm{xAUC}^1$ or $\mathrm{xAUC}^0$ Parity \cite{Kallus2019-yl,Cui2023-lo,Meissen2024-iu} &
\(\displaystyle \left|\mathrm{xAUC}^1_A-\mathrm{xAUC}^1_B\right|\)
\(\displaystyle \mathrm{or} \)
\(\displaystyle \left|\mathrm{xAUC}^0_A-\mathrm{xAUC}^0_B\right|\)
& Absolute difference in $\mathrm{xAUC}^1$ or $\mathrm{xAUC}^0$ across subgroups  & \textcolor{orange}{\textbf{Use with caution.}} Interpretation is difficult. Could be helpful when a single fairness metric is needed to summarise the differences observed in the AUC matrix.\\

\bottomrule
\end{tabular}
\end{table}

%% file: 6_euroscore.tex
The European System for Cardiac Operative Risk Evaluation (EuroSCORE) II \cite{nashef2012euroscore} is a clinical prediction model developed using data from  22,381 patients across 154 hospitals in 43 countries. It uses demographic and clinical predictors to estimate the probability of in-hospital mortality following cardiac surgery and achieved an AUC of 0.81. Although the model has been widely validated \cite{kalender2014cardiac,garcia2016validation,chalmers2013validation}, its fairness has been rarely assessed \cite{silverborn2023performance}. As a publicly-available model, EuroSCORE II may be used for, among others, three potential use cases, which we summarise below:

\begin{enumerate}
    \item \textbf{Risk estimation (continuous scale):} The model output can be used to communicate risk to patients and caregivers, benchmark institutional outcomes, or stratify populations' risk. 
    
    %Here, calibration is one of the most relevant performance domains because estimated probabilities should reflect true outcome frequencies. As with any risk model, its limitations must be recognised: the estimated mortality represents the average risk for a group of similar patients, not a definitive prediction for any one individual, who will ultimately either survive or die.
    
    \item \textbf{Binary decision:} A threshold (set by clinicians, hospitals, or health authorities) is applied, to facilitate surgical decision-making or serve as a quality control benchmark to stratify outcomes between low- and high-risk interventions. Decision thresholds are often arbitrarily set by optimising a statistical criterion.
    
    \item \textbf{Patient ranking:} In certain contexts, only the ordering of patients matters and the absolute magnitude of estimated risk is irrelevant. For example, a prioritisation system for surgical waiting lists might rank patients according to their estimated risk and relative ordering.
\end{enumerate}

% Consider a hypothetical scenario in which a new version of the EuroSCORE is developed from contemporary multicentre data, recalibrated to account for population drift and improvements in surgical outcomes. For the fairness evaluation, patient sex (at birth) is treated as a sensitive attribute. Sex is included in EuroSCORE II as a predictor, but the focus here is on assessing fairness in the model’s \textit{outputs}. 

In this section, we consider these three use cases as context for the interpretation of AUC-based fairness metrics, in three simulated performance scenarios. The Python code used to generate the simulated data, AUC-based fairness metrics, and figures presented in this section is publicly available as an executable Jupyter notebook on \href{https://colab.research.google.com/drive/12eu946PzBdCnz800UI0BwfZRziPTklON?usp=sharing}{Google Colab}.

\subsection{Simulations and discrimination performance scenarios}

We simulate three scenarios with varying discrimination performance across subgroups, defined by patient sex at birth (binary sensitive attribute). Simulations were conducted by generating estimated probabilities distributions separately for patients with and without events within each sex subgroup. For each scenario, we fixed the proportion of male and female patients ($p^M = 0.8$, $p^F = 0.2$) and the outcome prevalences ($\pi^M = 0.02$, $\pi^F = 0.03$), following the patient characteristics reported in \cite{silverborn2023performance}. Estimated probability distributions were simulated using truncated normal distributions with support on the interval $[0,1]$. Discrimination was controlled by specifying the means of the distributions patients not experiencing an event ($\mu_0^F$ and $\mu_0^M$) and experiencing an event ($\mu_1^F$ and $\mu_1^M$). Simulation parameters are summarised in Table~\ref{tab:scenario_parameters}. The resulting AUC metrics are reported in Table~\ref{tab:scenario_metrics}. Each scenario is examined below.

\begin{figure}[htbp]
\centering

\begin{subfigure}[t]{0.65\linewidth}
    \centering
    \includegraphics[width=\linewidth]{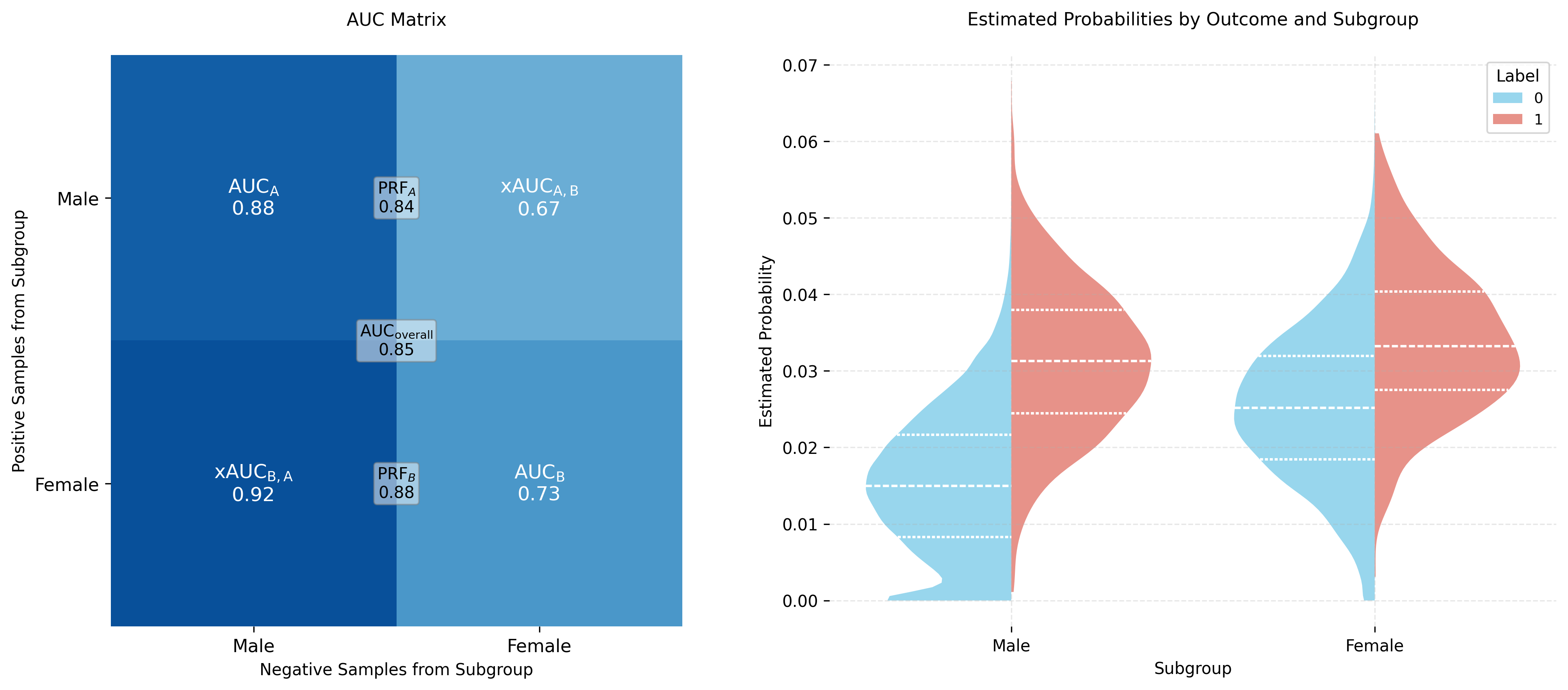}
    \caption{Simulated scenario illustrating substantial subgroup AUC disparity, reflecting an initial setup for the EuroSCORE case study. Estimated probability distributions follow $\mu_0^M = 0.015$, $\mu_1^M = 0.031$, $\mu_0^F = 0.025$, and $\mu_1^F = 0.034$. Labels of 0 denote patients without an event, labels of 1 denote patients with an event.}
    \label{fig:euro_fig1}
\end{subfigure}

\vspace{1em} % space between subfigures

\begin{subfigure}[t]{0.65\linewidth}
    \centering
    \includegraphics[width=\linewidth]{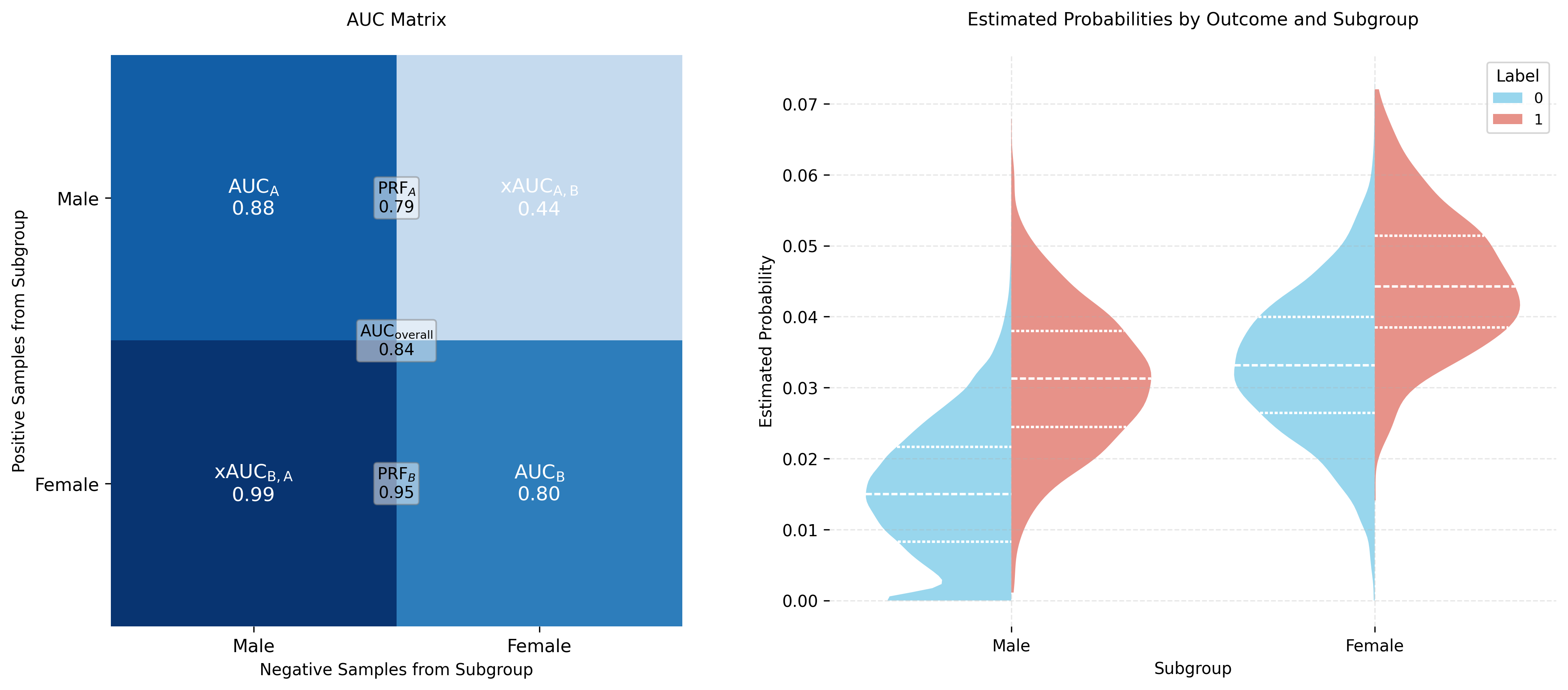}
    \caption{Simulated scenario illustrating a counter-intuitive ``level-up'' effect, in which improved subgroup performance does not change the overall AUC. Estimated probability distributions follow $\mu_0^M = 0.015$, $\mu_1^M = 0.031$, $\mu_0^F = 0.033$, and $\mu_1^F = 0.045$.}
    \label{fig:noncollap_example}

\end{subfigure}

\vspace{1em} % space between subfigures

\begin{subfigure}[t]{0.65\linewidth}
    \centering
    \includegraphics[width=\linewidth]{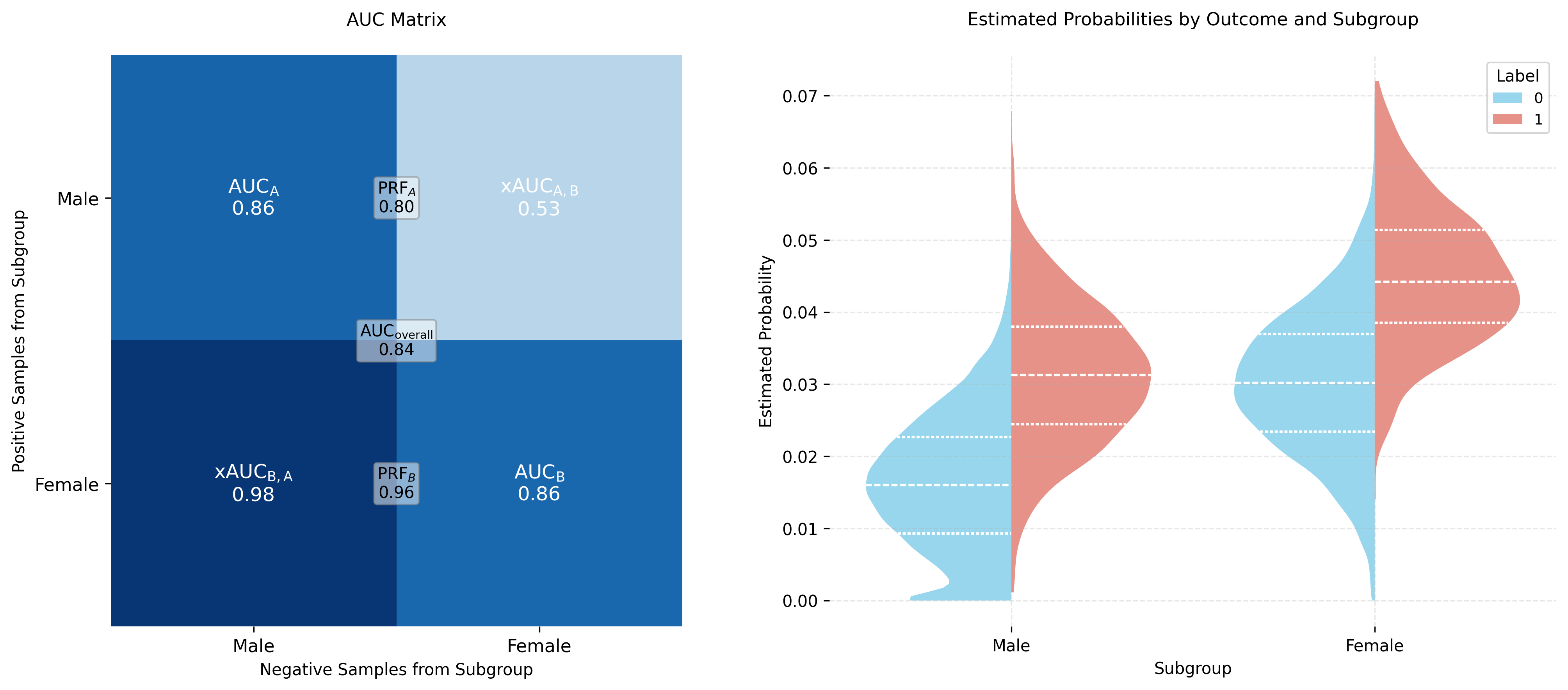}
    \caption{Simulated scenario in which subgroup parity is achieved ($\mathrm{AUC_A} = \mathrm{AUC_B}$) while the overall AUC falls outside the subgroup AUC range. Estimated probability distributions follow $\mu_0^M = 0.016$, $\mu_1^M = 0.031$, $\mu_0^F = 0.030$, and $\mu_1^F = 0.045$.}
    \label{fig:parity}

\end{subfigure}

\caption{\small Illustrative cases comparing subgroup and overall AUC relationships. Within each subgroup $g \in {M, F}$ (male, female, respectively) the estimated probabilities for non-events and events are drawn from truncated normal distributions on $[0,1]$, using 100,000 samples in total. For all three scenarios, $p^M = 0.8$, $\pi^M = 0.02$, $\pi^F = 0.03$, and the estimated probabilities follow $\sigma = 0.01$ for all strata. $\mu_0^M$, $\mu_1^M$, $\mu_0^F$, and $\mu_1^F$ are specific to each scenario. Labels of 0 denote patients without an event, labels of 1 denote patients with an event.}
\label{fig:combined_auc_euro}
\end{figure}

\paragraph{Scenario (i): Substantial within-group AUC disparity} 
Suppose the model’s overall AUC is 0.85, with within-group AUCs of 0.88 for male patients and 0.73 for female patients. However, there is a difference in cross-group AUC in the opposite direction, with \(\mathrm{xAUC}_{M,F} = 0.67\) and \(\mathrm{xAUC}_{F,M} = 0.92\). Metrics are summarised in Figure~\ref{fig:euro_fig1}.

\paragraph{Scenario (ii): Improving worst-performing subgroup performance}
The model is updated to improve discrimination in female patients, resulting in $\mathrm{AUC_{female}}$ increasing from 0.73 to 0.80, whilst performance in male patients remains unchanged ($\mathrm{AUC_{male}}$ = 0.88). Despite this improvement, the overall AUC decreases from 0.85 to 0.84, which is explained by a more pronounced disparity in cross-group AUC components: \(\mathrm{xAUC}_{M,F} = 0.44\) and \(\mathrm{xAUC}_{F,M} = 0.99\) (see Figure~\ref{fig:noncollap_example}).

\paragraph{Scenario (iii): Degrading best-performing subgroup performance} 
Suppose AUC parity is now achieved, such that the model appears fair from the standpoint of within-group AUC parity ($\mathrm{AUC_M} = \mathrm{AUC_F} = 0.86$). Improving \(\mathrm{AUC_F}\) came at the cost of degrading \(\mathrm{AUC_M}\). Despite within-group AUC parity, there still is a disparity in cross-group AUC: \(\mathrm{xAUC}_{M,F} = 0.53\) and \(\mathrm{xAUC}_{F,M} = 0.98\), which results in an overall AUC of 0.84 (see Figure~\ref{fig:parity}).

\begin{table}[htbp]
% \small
\centering
\caption{Simulation parameters. }
\label{tab:scenario_parameters}
\renewcommand{\arraystretch}{1.25}

\begin{tabular}{lcccccccc}
\hline
Scenario & $p^M$ & $\pi^M$ & $\pi^F$ &  $\mu_0^M$ & $\mu_1^M$ & $\mu_0^F$ & $\mu_1^F$ & $\sigma$\\
\hline
(i)   & 0.8 & 0.02 & 0.03 & 0.015 & 0.031 & 0.025 & 0.034 & 0.01\\
(ii)  & 0.8 & 0.02 & 0.03 & 0.015 & 0.031 & 0.033 & 0.045 & 0.01\\
(iii) & 0.8 & 0.02 & 0.03 & 0.016 & 0.031 & 0.030 & 0.045 & 0.01\\
\hline
\end{tabular}
\par
\vspace{0.5em}
\scriptsize
\textit{Legend:} $p^M$ denotes the proportion of male patients in the population. $\pi^M$ and $\pi^F$ are the outcome prevalence rates for male and female patients, respectively. 
$\mu_0^M$ and $\mu_1^M$ denotes mean of the (truncated normal) distribution of estimated probabilities for male patients without and with an event, respectively; $\mu_0^F$ and $\mu_1^F$ the mean of the distribution of estimated probabilities for female patients without and with an event. $\sigma$ denotes the standard deviation of each simulated normal distribution.
\end{table}

\begin{table}[ht]
\small
\centering
\caption{AUC metrics and AUC-based fairness metrics, across subgroups.}
\label{tab:scenario_metrics}
\renewcommand{\arraystretch}{1.25}

\begin{tabular}{lcccccccccccc}
\hline
Scenario
& $\mathrm{AUC}_{\mathrm{global}}$
& $\mathrm{AUC}_{M}$
& $\mathrm{AUC}_{F}$
& $\mathrm{xAUC}_{M,F}$
& $\mathrm{xAUC}_{F,M}$
& $\mathrm{xAUC}^1_{M}$
& $\mathrm{xAUC}^1_{F}$
& $\mathrm{AUC}_{\mathrm{parity}}$
& $\mathrm{xAUC}_{\mathrm{parity}}$
& $\mathrm{xAUC}^1_{\mathrm{parity}}$\\
\hline
(i)   & 0.85 & 0.88 & 0.73 & 0.67 & 0.92 & 0.84 & 0.88 & 0.15 & 0.25 & 0.04\\
(ii)  & 0.84 & 0.88 & 0.80 & 0.44 & 0.99 & 0.79 & 0.95 & 0.08 & 0.55 & 0.16\\
(iii) & 0.84 & 0.86 & 0.86 & 0.53 & 0.98 & 0.80 & 0.96 & 0.00 & 0.45 & 0.16\\
\hline
\end{tabular}
\end{table}

\subsection{Fairness interpretation across use cases and scenarios}

\subsubsection{Model use case (1): Risk estimation}

Let us assume that the developed model is used to communicate risk. The model estimated probabilities are directly used, and no decision thresholds are applied. We consider three scenarios.

% message 1: within-group disparity can co-exist with cross-group disparity in the opposite direction - the worse-performing group may be doing better
\paragraph{Scenario (i): Substantial AUC disparity}

A difference in within-group AUCs exists ($\mathrm{AUC}_{\mathrm{parity}} = \left | 0.88 - 0.73 \right | = 0.15$). However, when examining $\mathrm{xAUC}$ components, the picture differs: $\mathrm{xAUC}_{M,F} = 0.67$ and $\mathrm{xAUC}_{F,M} = 0.92$, which means that female patients with an event would be better separated from male patients without an event ($\mathrm{xAUC}_{F,M}$), than the other other way around ($\mathrm{xAUC}_{M,F}$), resulting in a cross-group disparity of 0.25 ($\mathrm{xAUC}_{\mathrm{parity}} = \left | 0.67 - 0.92 \right | = 0.25$). Here, $\mathrm{AUC}_{\mathrm{parity}}$ does not provide a complete picture of unfairness, and $\mathrm{xAUC}_{\mathrm{parity}}$ is informative.

When examining $\mathrm{xAUC}^1$ metrics, the interplay between within- and cross-group discrimination is summarised: \(\mathrm{xAUC}^1_M = 0.84\) and \(\mathrm{xAUC}^1_F = 0.88\), resulting in a less pronounced disparity according to these metrics: \(\mathrm{xAUC}^1_{\mathrm{parity}} = \left | 0.84-0.88 \right | = 0.04\). This results from the fact that male patients are better off according to within-group AUCs, whereas female patients are better off in terms of cross-group AUC. Here, \(\mathrm{xAUC}^1_{\mathrm{parity}}\) is not exactly zero, meaning that the model is not considered ``perfectly fair'' according to this metric, which happens because the ``advantage'' that female patients have in cross-group discrimination outweighs the ``advantage'' that male patients have in within-group discrimination.

\textbf{Key takeaways:} Within-group AUC disparity can co-exist with cross-group AUC disparity, underscoring the importance of reporting both within-group AUC parity and cross-group AUC parity metrics.

% message 2: improving the performance of the worse-performing group in terms of within-group AUC may result in a lower overall AUC if the cross-group AUCs move in greater magnitude in the opposite direction
\paragraph{Scenario (ii): Improving worst-performing subgroup performance}

In this new scenario, the difference in within-group AUC has decreased, thanks to increase in \(\mathrm{AUC}_F\) from 0.73 to 0.80, while \(\mathrm{AUC}_M\) was kept at 0.88. As depicted on the right-hand side of Figure~\ref{fig:noncollap_example}, this is explained by the fact that the distribution of model estimated probabilities for male patients did not change, only female patients' distribution did. Nonetheless, this resulted in changes in cross-group AUC values that affected both subgroups' performance. \(\mathrm{xAUC}_{M,F}\) further deteriorated, decreasing from 0.67, in Scenario (i), to 0.44 here, whereas \(\mathrm{xAUC}_{F,M}\) improved, going from 0.92, in Scenario (i), to 0.99 here. As a result, overall AUC degraded slightly, going from 0.85 in Scenario (i), to 0.84, meaning that the model discrimination improvement in terms of \(\mathrm{AUC}_F\) did not compensate for losses in cross-group discrimination (in which both \(\mathrm{xAUC}_{M,F}\) and \(\mathrm{xAUC}_{F,M}\) changed).

As a result of these model changes, \(\mathrm{xAUC}^1_M\) declines from 0.84 in Scenario (i) to 0.79 here, highlighting the contribution of both \(\mathrm{AUC}_M\) and \(\mathrm{xAUC}_{M,F}\) to \(\mathrm{xAUC}^1_M\). The verified decline in \(\mathrm{xAUC}^1_M\) also shows that, even though the model estimated probabilities for male patients did not change, discrimination declined for these patients, overall, due to the contribution of cross-group ranking. On the other hand, \(\mathrm{xAUC}^1_F\) improved, going from 0.88, in Scenario (i), to 0.95 in this Scenario (ii), which stems from the improved \(\mathrm{xAUC}_{F,M}\) and \(\mathrm{AUC}_{F}\).

In terms of fairness metrics, the improved \(\mathrm{AUC}_{F}\) results in a lower \(\mathrm{AUC}_{\mathrm{parity}} = \left | 0.88 - 0.80 \right | = 0.08\), meaning that the model in Scenario (ii) is fairer than the model in Scenario (i) in terms of within-group discrimination. However, \(\mathrm{xAUC}_{parity} = \left | 0.44 - 0.99 \right | = 0.55\), which is substantially higher than the 0.25 disparity in cross-group discrimination from Scenario (i), meaning that this new model is more unequal in terms of cross-group discrimination, to a greater extent than the fairness improvement in within-group discrimination. The \(\mathrm{xAUC}^1\) metrics show this: \(\mathrm{xAUC}^1_{\mathrm{parity}} = \left | 0.79 - 0.95 \right | = 0.16\), an increase in disparity when compared to Scenario (i) (0.04). This parity metric indicates greater inequality across male and female patients, when taking into consideration both within- and cross-group discrimination.

An important artefact in this Scenario (ii) is the apparent improvement in ``subgroup AUCs'' (i.e., improvement of female patients' within-group AUC), but a decline in overall AUC. This almost paradoxical finding is a consequence of the AUC non-collapsibility \cite{matos2026collapsibility}, and highlights the importance of considering and reporting cross-group AUCs. The phenomena we describe here complicate interpretation and communication with stakeholders. What was intended to be a ``level-up'' strategy, improving performance (measured in terms of within-group AUC) of the worse-off subgroup (female patients), without compromising the better-off subgroup (male patients), inadvertently produced a ``level-down'' effect for male patients through changes in cross-group discrimination.

\textbf{Key takeaways:} Improving the performance of a worse-performing subgroup in terms of within-group AUC, without any model change to the model estimated probabilities of the better-performing subgroup, may nonetheless result in a deterioration of the unchanged subgroup's cross-group AUC. If such deterioration is in greater magnitude than the within-group improvement, a (seemingly paradoxical) degradation of overall AUC can occur.

% message 3: when parity in within-group is achieved, it can co-exist with disparity in cross-group disparity, and if that assymetry is pronounced, it can even mean that the overall AUC is lower than both subgroups' within-group AUC
\paragraph{Scenario (iii): Degrading best-performing subgroup performance}

In this final scenario, within-group AUC parity is achieved (\(\mathrm{AUC}_{\mathrm{parity}} = \left | 0.86 - 0.86 \right | = 0\)) thanks to a model modification that shifts the estimated probabilities distributions of both male and female patients (in Scenario (ii) only female patients' distribution had been affected). As a result, \(\mathrm{AUC}_M\) declines to 0.86, from 0.88 in Scenario (i) and (ii), equalising \(\mathrm{AUC}_F\) at 0.86, which raises from 0.80 in Scenario (ii). This model modification also results in an increase of \(\mathrm{xAUC}_{M,F}\) from 0.44 in Scenario (ii) to 0.53 here, and a slight decrease in \(\mathrm{xAUC}_{F,M}\) from 0.99 in Scenario (ii) to 0.98 here, making \(\mathrm{xAUC}_{\mathrm{parity}} = \left | 0.53 - 0.98 \right | = 0.45\). Overall AUC is kept at 0.84, which is below both \(\mathrm{AUC}_M = \mathrm{AUC}_F = 0.86 \), an ``artefact'' of AUC's non-collapsibility \cite{matos2026collapsibility} that is possible here due to cross-group AUC values (\(\mathrm{xAUC}_{M,F}=0.53\) and \(\mathrm{xAUC}_{F,M}=0.98\)), which together outweigh the within-group values (\(\mathrm{AUC}_{M}=\mathrm{AUC}_F=0.86\)).

Additionally, \(\mathrm{xAUC}^1_M\) increased from 0.79 in Scenario (ii) to 0.80 here, and \(\mathrm{xAUC}^1_F\) from 0.95 in Scenario (ii) to 0.96 here, resulting in \(\mathrm{xAUC}^1_{\mathrm{parity}} = 0.16\), the same exact value achieved in Scenario (ii). The fact that \(\mathrm{xAUC}^1_{\mathrm{parity}}\) did not change from Scenario (ii) to (iii) underscores the importance of considering both within- and cross-group metrics to understand sources of potential unfairness. Despite the apparent deterioration captured by \(\mathrm{AUC}_M\) (\(\mathrm{AUC}_M = 0.88\) in Scenario (ii), to \(\mathrm{AUC}_M = 0.86\) here), male patients achieve better performance in terms of model discrimination, overall, because of the cross-group AUC contributions (\(\mathrm{xAUC}_{M,F} = 0.44\) in Scenario (ii), raised to \(\mathrm{xAUC}_{M,F} = 0.53\) here).

\textbf{Key takeaways:} Within-group AUC parity can co-exist with cross-group AUC disparity. If cross-group AUC disparity is sufficiently large, the resulting overall AUC could be lower (or higher) than both subgroups' within-group AUC values. This artefact is created by AUC non-collapsibility and may complicate communication with stakeholders.

\subsubsection{Model use case (2): Binary decision}

Let us assume now that this last model -- \textit{Scenario (iii)} -- is used for binary decisions, i.e., a threshold is applied to model estimated probabilities. Two plausible \emph{strategies} could follow.

\paragraph{Strategy (i): One single threshold is applied to all patients.} 

When a model is used to support decision-making, all patients (regardless of their subgroups) use a common decision threshold. With such strategy, relying on within-group discrimination metrics ($\mathrm{AUC_g}$) alone is insufficient. Cross-group AUC terms are as important as within-group AUC values because together they determine how estimated probabilities are ordered when a single threshold is applied.

In this Scenario (iii) model, as shown in Figure \ref{fig:parity}, even when within-group AUCs are identical (\(\mathrm{AUC}_M = \mathrm{AUC}_F = 0.86\)), \(\mathrm{xAUC}_{M,F} = 0.53\) versus \(\mathrm{xAUC}_{F,M} = 0.98\) indicate that it is only slightly above that chance (53\% probability) that male patients with an event are assigned estimated probabilities that are higher than the ones assigned to female patients without an event, whereas female patients with an event will have a higher estimated probability than male patients without an event most times (98\% probability). In settings where the decision is polar (i.e., patients benefit from selection \cite{Paulus_Kent_2020}), this asymmetry disadvantages male patients, as female patients are far more likely to exceed the decision threshold. With a single decision threshold, this asymmetry would result in unequal downstream decisions (which could be captured by threshold-dependent fairness metrics \cite{matos2025critical}).

Two main mechanisms can explain how within-group AUC parity can coexist with cross-group AUC disparities. First, differences in outcome prevalence across subgroups may reflect differences in underlying risk, leading to shifts in the distributions of estimated probabilities. For example, female patients have a higher outcome prevalence (\(\pi^F = 0.03\)) than male patients (\(\pi^M = 0.02\)), which may result in systematically higher estimated probabilities. Such differences can be monitored using ``mean score parity'' metrics \cite{matos2026collapsibility,matos2025critical}. Second, differences in calibration across subgroups can lead to asymmetric cross-group rankings, even when within-group AUCs are identical. For instance, if risk is systematically underestimated for male patients but not for female patients, the resulting differences in estimated probabilities may introduce cross-group AUC disparities \cite{matos2026collapsibility}.

\textbf{Key takeaways:} Applying a single decision threshold to a model with \(\mathrm{AUC}_{\mathrm{parity}} = 0\) but \(\mathrm{xAUC}_{\mathrm{parity}} \gg 0\) is expected to result in unequal downstream decisions, underscoring the importance of reporting within- and cross-group discrimination to prevent downstream unequal decisions.

\paragraph{Strategy (ii): Subgroup specific thresholds are defined and applied.}

When cross-group AUC disparities exist (\(\mathrm{xAUC}_{\mathrm{parity}} = 0.45\)) but no within-group AUC disparities exist (\(\mathrm{AUC}_{\mathrm{parity}} = 0\)), as seen in Scenario (iii), subgroup specific thresholds should be considered. If distinct decision thresholds are specified for each subgroup, cross-group comparisons no longer directly influence decision-making and the within-group AUC parity becomes the only consequential fairness metric in this context. Fairness considerations in terms of discrimination only concern within-group AUCs, as cross-group ``competition'' is eliminated and cross-group ranking ceases to be a consideration. Subgroup specific thresholds (such as ``race corrections'' \cite{vyas2020hidden}) are not uncommon in medicine. 

\textbf{Key takeaways:} Subgroup specific decision thresholds should be considered when \(\mathrm{xAUC}_{\mathrm{parity}} \gg 0\). With subgroup specific decision thresholds, cross-group ranking ceases to be a consideration, and fairness in terms of discrimination solely depends on \(\mathrm{AUC}_{\mathrm{parity}}\).

\subsubsection{Model use case (3): Patient ranking}

Finally, let us assume that the model is being used to rank patients. In settings where only the ordering of individuals matters, such as surgery prioritisation, absolute risk magnitudes become irrelevant. Decisions in this context are inherently polar, as patients are ``competing'' against each other (and against the ``system'') to get an estimated probability that places them on top of the list \cite{Paulus_Kent_2020}. Traditional within-group AUC parity-based fairness analyses are insufficient to assess ranking/discrimination fairness, and xAUC components are equally relevant. Metrics that combine both within- and cross-group AUCs, such as the $\mathrm{xAUC}^1$ and $\mathrm{xAUC}^0$, are therefore well suited to summarise fairness in terms of discrimination in this setting (Figure \ref{fig:combined_auc_euro}). 

%% file: 7_conclusion.tex
In this paper, we examined eight AUC-based fairness metrics, characterised their estimands, critically appraised their interpretation using a clinical example, and provided guidance on how to use and interpret these metrics in the context of different types of fairness assessments. The use and interpretation of these metrics are inherently context-dependent, and no single metric can fully capture the complexity of fairness in model discrimination when subgroups coexist. Therefore, we recommend reporting the AUC matrix as a whole, containing the overall AUC and all within- and cross-group AUC components, to allow discrimination performance to be evaluated comprehensively. When fairness metrics are needed (for example, when more than two subgroups coexist, resulting in large AUC matrices, or when a summary measure of fairness is preferable), we recommend within-group AUC parity as the primary fairness metric for discrimination performance. Cross-group AUC parity can provide important additional information on cross-group ranking disparities but should be used with caution due to its potential conflict with calibration and case-mix differences across subgroups. Transparent reporting of population characteristics \cite{matos2026collapsibility} and adherence to TRIPOD+AI \cite{collins2024tripod+} are essential for meaningful fairness evaluations of clinical predictive AI. Although this paper focused on discrimination and AUC-based fairness metrics, their interplay with calibration is important and warrants further investigation, especially regarding the potential trade-offs between discrimination and calibration fairness.

%% file: appendix.tex
Starting from the overall AUC,

\[
  \mathrm{AUC} = P(\hat p_{1}>\hat p_{0})
  = \sum_{i\in\{A,B\}}\sum_{j\in\{A,B\}}
    P(g=i\mid Y=1)\,P(g=j\mid Y=0)\,
    P(\hat p_{1}^{i}>\hat p_{0}^{j}).
\]

Collapsing over the negative group yields $\mathrm{xAUC}^1$:

\begin{align}
  \mathrm{AUC}
  &= \sum_{i\in\{A,B\}} P(g=i\mid Y=1)
      \sum_{j\in\{A,B\}} P(g=j\mid Y=0)\,P(\hat p_{1}^{i}>\hat p_{0}^{j}) \notag\\
  &= \sum_{i\in\{A,B\}} P(g=i\mid Y=1)\,P(\hat p_{1}^{i}>\hat p_{0}\notag)\\
  &= \sum_{i\in\{A,B\}} P(g=i\mid Y=1)\,\mathrm{xAUC}^1_{i}. \label{eq:auc-prf}
\end{align}

Collapsing over the positive group yields $\mathrm{xAUC}^0$:

\begin{align}
  \mathrm{AUC}
  &= \sum_{j\in\{A,B\}} P(g=j\mid Y=0)
      \sum_{i\in\{A,B\}} P(g=i\mid Y=1)\,P(\hat p_{0}^{j}<\hat p_{1}^{i}) \notag\\
  &= \sum_{j\in\{A,B\}} P(g=j\mid Y=0)\,P(\hat p_{1}>\hat p_{0}^{j})\notag\\
  &= \sum_{j\in\{A,B\}} P(g=j\mid Y=0)\,\mathrm{xAUC}^0_{j}. \label{eq:auc-sauc}
\end{align}

Substituting \(P(g=i\mid Y=1)=p_{1}^{i}/p_{1}\) and \(P(g=j\mid Y=0)=p_{0}^{j}/p_{0}\) gives the concise decomposition:

\begin{align}
    \mathrm{AUC}
    = \frac{p_{1}^{A}}{p_{1}}\,\mathrm{xAUC}^1_{A}
      + \frac{p_{1}^{B}}{p_{1}}\,\mathrm{xAUC}^1_{B}
    = \frac{p_{0}^{A}}{p_{0}}\,\mathrm{xAUC}^0_{A}
      + \frac{p_{0}^{B}}{p_{0}}\,\mathrm{xAUC}^0_{B}
   \label{eq:auc-prf-sauc}
\end{align}

Comparing Equation \ref{eq:auc_decomposition_two_groups} with the $\mathrm{xAUC}^1$ decomposition \eqref{eq:auc-prf} yields

\begin{align}
  \mathrm{xAUC}^1_{A} = \frac{p_{0}^{A}}{p_{0}}\,\mathrm{AUC}_{A}
                  + \frac{p_{0}^{B}}{p_{0}}\,\mathrm{xAUC}_{A,B}
    \label{prf_a}
\end{align}

\begin{align}
  \mathrm{xAUC}^1_{B} = \frac{p_{0}^{A}}{p_{0}}\,\mathrm{xAUC}_{B,A}
                  + \frac{p_{0}^{B}}{p_{0}}\,\mathrm{AUC}_{B}
                  \label{prf_b}
\end{align}

and, similarly, for $\mathrm{xAUC}^0$ decomposition \ref{eq:auc-sauc}:

\begin{align}
 \mathrm{xAUC}^0_{A} = \frac{p_{1}^{A}}{p_{1}}\,\mathrm{AUC}_{A}
                 + \frac{p_{1}^{B}}{p_{1}}\,\mathrm{xAUC}_{B,A}
\label{sauc_a}
\end{align}

\begin{align}
  \mathrm{xAUC}^0_{B} = \frac{p_{1}^{A}}{p_{1}}\,\mathrm{xAUC}_{A,B}
                 + \frac{p_{1}^{B}}{p_{1}}\,\mathrm{AUC}_{B}
\label{sauc_b}
\end{align}

%% file: sample-base.bib
@String{Springer = "Springer-Verlag" }

@ARTICLE{Kallus2019-yl,
  title   = "The fairness of risk scores beyond classification: Bipartite
             ranking and the xauc metric",
  author  = "Kallus, Nathan and Zhou, Angela",
  journal = "Advances in neural information processing systems",
  volume  =  32,
  year    =  2019,
  doi     = "https://doi.org/10.48550/arXiv.1902.05826"
}

@ARTICLE{Luo2024-uo,
  title     = "Harvard glaucoma fairness: a retinal nerve disease dataset for
               fairness learning and fair identity normalization",
  author    = "Luo, Yan and Tian, Yu and Shi, Min and Pasquale, Louis R and
               Shen, Lucy Q and Zebardast, Nazlee and Elze, Tobias and Wang,
               Mengyu",
  journal   = "IEEE Transactions on Medical Imaging",
  publisher = "IEEE",
  year      =  2024,
  doi       = "https://doi.org/10.1109/TMI.2024.3377552"
}

@ARTICLE{Cui2023-lo,
  title     = "Bipartite ranking fairness through a model agnostic ordering
               adjustment",
  author    = "Cui, Sen and Pan, Weishen and Zhang, Changshui and Wang, Fei",
  journal   = "IEEE Transactions on Pattern Analysis and Machine Intelligence",
  publisher = "IEEE",
  year      =  2023,
  doi       = "https://doi.org/10.48550/arXiv.2307.14668"
}

@ARTICLE{Meissen2024-iu,
  title     = "(Predictable) performance bias in unsupervised anomaly detection",
  author    = "Meissen, Felix and Breuer, Svenja and Knolle, Moritz and Buyx,
               Alena and Müller, Ruth and Kaissis, Georgios and Wiestler,
               Benedikt and Rückert, Daniel",
  journal   = "Ebiomedicine",
  publisher = "Elsevier",
  volume    =  101,
  year      =  2024,
  doi       = "https://doi.org/10.1016/j.ebiom.2024.105002"
}

@ARTICLE{Van-Calster2024-jw,
  title         = "Evaluation of performance measures in predictive artificial intelligence models to support medical decisions: overview and guidance",
  author        = "Van Calster, Ben and Collins, Gary S and Vickers, Andrew J and Wynants, Laure and Kerr, Kathleen F and Barre{\~n}ada, Lasai and Varoquaux, Gael and Singh, Karandeep and Moons, Karel GM and Hernandez-Boussard, Tina and others",
  journal       = "The Lancet Digital Health",
  publisher = "Elsevier",
  month         =  dec,
  year          =  2025,
  doi           = "https://doi.org/10.1016/j.landig.2025.100916"
}

@article{matos2025critical,
  title={Critical Appraisal of Fairness Metrics in Clinical Predictive AI},
  author={Matos, Jo{\~a}o and Van Calster, Ben and Celi, Leo Anthony and Dhiman, Paula and Gichoya, Judy Wawira and Riley, Richard D and Russell, Chris and Khalid, Sara and Collins, Gary S},
  journal={The Lancet Digital Health},
  year={2026},
  doi={https://doi.org/10.1016/j.landig.2026.101001}
}

@article{kalender2014cardiac,
  title={Validation of EuroSCORE II risk model for coronary artery bypass surgery in high-risk patients},
  author={Kalender, Mehmet and Adademir, Taylan and Tasar, Mehmet and Ecevit, Ata Niyazi and Karaca, Okay Guven and Salihi, Salih and Buyukbayrak, Fuat and Ozkokeli, Mehmet},
  journal={Kardiochirurgia i Torakochirurgia Polska/Polish Journal of Thoracic and Cardiovascular Surgery},
  volume={11},
  number={3},
  pages={252--256},
  year={2014},
  publisher={Termedia},
  url={https://pubmed.ncbi.nlm.nih.gov/26336431/}
}

@article{garcia2016validation,
  title={Validation and quality measurements for EuroSCORE and EuroSCORE II in the Spanish cardiac surgical population: a prospective, multicentre study},
  author={Garcia-Valentin, Antonio and Mestres, Carlos A and Bernabeu, Eduardo and Bahamonde, Jos{\'e} A and Mart{\'\i}n, Iv{\'a}n and Rueda, Cristina and Domenech, Alberto and Valencia, Jamit and Fletcher, Delfina and Machado, Facundo and others},
  journal={European Journal of Cardio-Thoracic Surgery},
  volume={49},
  number={2},
  pages={399--405},
  year={2016},
  publisher={Oxford University Press},
  doi={https://doi.org/10.1093/ejcts/ezv090}
}

@article{chalmers2013validation,
  title={Validation of EuroSCORE II in a modern cohort of patients undergoing cardiac surgery},
  author={Chalmers, John and Pullan, Mark and Fabri, Brian and McShane, James and Shaw, Matthew and Mediratta, Neeraj and Poullis, Michael},
  journal={European Journal of Cardio-Thoracic Surgery},
  volume={43},
  number={4},
  pages={688--694},
  year={2013},
  publisher={Oxford University Press},
  doi={https://doi.org/10.1093/ejcts/ezs406}
}

@article{nashef2012euroscore,
  title={Euroscore ii},
  author={Nashef, Samer AM and Roques, Fran{\c{c}}ois and Sharples, Linda D and Nilsson, Johan and Smith, Christopher and Goldstone, Antony R and Lockowandt, Ulf},
  journal={European journal of cardio-thoracic surgery},
  volume={41},
  number={4},
  pages={734--745},
  year={2012},
  publisher={Oxford University Press},
  doi = {https://doi.org/10.1093/ejcts/ezs043}
}

@article{silverborn2023performance,
  title={The performance of EuroSCORE II in CABG patients in relation to sex, age, and surgical risk: a nationwide study in 14,118 patients},
  author={Silverborn, Martin and Nielsen, Susanne and Karlsson, Martin},
  journal={Journal of Cardiothoracic Surgery},
  volume={18},
  number={1},
  pages={40},
  year={2023},
  publisher={Springer},
  doi={https://doi.org/10.1186/s13019-023-02141-4}
}

@article{Paulus_Kent_2020,
    title={Predictably unequal: Understanding and addressing concerns that algorithmic clinical prediction may increase health disparities},
    volume={3},
    DOI={10.1038/s41746-020-0304-9},
    number={1},
    journal={npj Digital Medicine},
    author={Paulus, Jessica K. and Kent, David M.},
    year={2020},
    month={Jul}
}

@article{mittelstadt2023unfairnessfairmachinelearning,
      title={The Unfairness of Fair Machine Learning: Levelling down and strict egalitarianism by default}, 
      author={Brent Mittelstadt and Sandra Wachter and Chris Russell},
      year={2023},
      eprint={2302.02404},
      archivePrefix={arXiv},
      primaryClass={cs.AI},
      url={https://arxiv.org/abs/2302.02404}, 
}

@article{byrd2023validation,
  title={Validation of a proprietary deterioration index model and performance in hospitalized adults},
  author={Byrd, Thomas F and Southwell, Bronwyn and Ravishankar, Adarsh and Tran, Travis and Kc, Abhinab and Phelan, Tom and Melton-Meaux, Genevieve B and Usher, Michael G and Scheppmann, Daren and Switzer, Sean and others},
  journal={JAMA Network Open},
  volume={6},
  number={7},
  pages={e2324176--e2324176},
  year={2023},
  publisher={American Medical Association},
  doi={https://doi.org/10.1001/jamanetworkopen.2023.24176}
}

@article{albuquerque2024evaluating,
  title={Evaluating model bias requires characterizing its mistakes},
  author={Albuquerque, Isabela and Schrouff, Jessica and Warde-Farley, David and Cemgil, Taylan and Gowal, Sven and Wiles, Olivia},
  journal={arXiv preprint arXiv:2407.10633},
  year={2024},
  doi={https://doi.org/10.48550/arXiv.2407.10633}
}

@article{janssens2020reflection,
  title={Reflection on modern methods: Revisiting the area under the ROC Curve},
  author={Janssens, A Cecile JW and Martens, Forike K},
  journal={International journal of epidemiology},
  volume={49},
  number={4},
  pages={1397--1403},
  year={2020},
  publisher={Oxford University Press},
  doi={https://doi.org/10.1093/ije/dyz274}
}

@article{collins2024tripod+,
  title={TRIPOD+ AI statement: updated guidance for reporting clinical prediction models that use regression or machine learning methods},
  author={Collins, Gary S and Moons, Karel GM and Dhiman, Paula and Riley, Richard D and Beam, Andrew L and Van Calster, Ben and Ghassemi, Marzyeh and Liu, Xiaoxuan and Reitsma, Johannes B and Van Smeden, Maarten and others},
  journal={bmj},
  volume={385},
  year={2024},
  publisher={British Medical Journal Publishing Group},
  doi={https://doi.org/10.1136/bmj-2023-078378}
}

@inproceedings{diana2021minimax,
  title={Minimax group fairness: Algorithms and experiments},
  author={Diana, Emily and Gill, Wesley and Kearns, Michael and Kenthapadi, Krishnaram and Roth, Aaron},
  booktitle={Proceedings of the 2021 AAAI/ACM Conference on AI, Ethics, and Society},
  pages={66--76},
  year={2021},
  doi={https://doi.org/10.1145/3461702.3462523}
}

@article{vyas2020hidden,
  title={Hidden in plain sight—reconsidering the use of race correction in clinical algorithms},
  author={Vyas, Darshali A and Eisenstein, Leo G and Jones, David S},
  journal={New England Journal of Medicine},
  volume={383},
  number={9},
  pages={874--882},
  year={2020},
  publisher={Mass Medical Soc},
  doi={10.1056/NEJMms2004740}
}

@article{collins2024evaluation,
  title={Evaluation of clinical prediction models (part 1): from development to external validation},
  author={Collins, Gary S and Dhiman, Paula and Ma, Jie and Schlussel, Michael M and Archer, Lucinda and Van Calster, Ben and Harrell, Frank E and Martin, Glen P and Moons, Karel GM and Van Smeden, Maarten and others},
  journal={bmj},
  volume={384},
  year={2024},
  publisher={British Medical Journal Publishing Group},
  doi={https://doi.org/10.1136/bmj-2023-074819}
}

@article{topol2019high,
  title={High-performance medicine: the convergence of human and artificial intelligence},
  author={Topol, Eric J},
  journal={Nature medicine},
  volume={25},
  number={1},
  pages={44--56},
  year={2019},
  publisher={Nature Publishing Group US New York},
  doi={https://doi.org/10.1038/s41591-018-0300-7}
}

@article{mccradden2020ethical,
  title={Ethical limitations of algorithmic fairness solutions in health care machine learning},
  author={McCradden, Melissa D and Joshi, Shalmali and Mazwi, Mjaye and Anderson, James A},
  journal={The Lancet Digital Health},
  volume={2},
  number={5},
  pages={e221--e223},
  year={2020},
  publisher={Elsevier},
  doi="https://doi.org/10.1016/s2589-7500(20)30065-0"
}

@article{Oakden-Rayner2020-fn,
	title        = {Hidden stratification causes clinically meaningful failures in machine learning for medical imaging},
	author       = {Oakden-Rayner, Luke and Dunnmon, Jared and Carneiro, Gustavo and R\'{e}, Christopher},
	year         = 2020,
	month        = apr,
	journal      = {Proc. ACM Conf. Health Inference Learn.},
	volume       = 2020,
	pages        = {151--159},
	language     = {en},
    doi={https://doi.org/10.1145/3368555.3384468}
    
}

@ARTICLE{Pencina2015-uu,
  title     = "Evaluating discrimination of risk prediction models: The {C}
               statistic",
  author    = "Pencina, Michael J and D'Agostino, Sr, Ralph B",
  journal   = "JAMA",
  publisher = "American Medical Association (AMA)",
  volume    =  314,
  number    =  10,
  pages     =  1063,
  month     =  sep,
  year      =  2015,
  doi       = "http://doi.org/10.1001/jama.2015.11082"
}

@ARTICLE{matos2026collapsibility,
      title={Collapsibility of Performance Metrics in Clinical Predictive AI}, 
      author={João Matos and Ben Van Calster and Richard D. Riley and Paula Dhiman and Gary S. Collins},
      year={2026},
      journal={arXiv preprint arXiv:2608.30568},
      archivePrefix={arXiv},
      primaryClass={cs.LG},
      doi={https://doi.org/10.48550/arXiv.2608.30568}, 
}

@inproceedings{chang2022disparate,
  title={Disparate censorship \& undertesting: A source of label bias in clinical machine learning},
  author={Chang, Trenton and Sjoding, Michael W and Wiens, Jenna},
  booktitle={Machine Learning for Healthcare Conference},
  pages={343--390},
  year={2022},
  organization={PMLR},
  url={https://pubmed.ncbi.nlm.nih.gov/37152303/}
}

@inproceedings{cheng2021can,
  title={Can you fake it until you make it? impacts of differentially private synthetic data on downstream classification fairness},
  author={Cheng, Victoria and Suriyakumar, Vinith M and Dullerud, Natalie and Joshi, Shalmali and Ghassemi, Marzyeh},
  booktitle={Proceedings of the 2021 ACM Conference on Fairness, Accountability, and Transparency},
  pages={149--160},
  year={2021},
  doi={https://doi.org/10.1145/3442188.3445879}
}

@article{gallifant2024ethical,
  title={Ethical debates amidst flawed healthcare artificial intelligence metrics},
  author={Gallifant, Jack and Bitterman, Danielle S and Celi, Leo Anthony and Gichoya, Judy W and Matos, Joao and McCoy, Liam G and Pierce, Robin L},
  journal={NPJ Digital Medicine},
  volume={7},
  number={1},
  pages={243},
  year={2024},
  publisher={Nature Publishing Group UK, London},
  doi={https://doi.org/10.1038/s41746-024-01242-1}
}

@article{matos2026jama,
    author = {Matos, João and Dhiman, Paula and Collins, Gary S.},
    title = {Clarifying the SOFA-2 Score},
    journal = {JAMA},
    volume = {335},
    number = {15},
    pages = {1359-1360},
    year = {2026},
    month = {04},
    issn = {0098-7484},
    doi = {10.1001/jama.2026.0310},
    url = {https://doi.org/10.1001/jama.2026.0310},
}

@article{matos2024clinician,
  title={A clinician’s guide to understanding bias in critical clinical prediction models},
  author={Matos, Jo{\~a}o and Gallifant, Jack and Chowdhury, Anand and Economou-Zavlanos, Nicoleta and Charpignon, Marie-Laure and Gichoya, Judy and Celi, Leo Anthony and Nazer, Lama and King, Heather and Wong, An-Kwok Ian},
  journal={Critical Care Clinics},
  volume={40},
  number={4},
  pages={827--857},
  year={2024},
  publisher={Elsevier},
  doi={https://doi.org/10.1016/j.ccc.2024.05.011}
}
